# Optimising Fault-Tolerant Quality-Guaranteed Sensor Deployments for UAV Localisation in Critical Areas via Computational Geometry

Marco Esposito, Toni Mancini, and Enrico Tronci

***Abstract*—The increasing spreading of small commercial Unmanned Aerial Vehicles (UAVs, aka *drones*) presents serious threats for critical areas such as airports, power plants, governmental and military facilities. In fact, such UAVs can easily disturb or jam radio communications, collide with other flying objects, perform espionage activity, and carry offensive payloads, *e.g.*, weapons or explosives. A central problem when designing surveillance solutions for the *localisation* of unauthorised UAVs in critical areas is to decide how many triangulating sensors to use, and where to deploy them to optimise both *coverage* and *cost effectiveness*.**

**In this article, we compute deployments of triangulating sensors for UAV localisation, optimising a given blend of metrics, namely: coverage under multiple sensing quality levels, cost-effectiveness, fault-tolerance. We focus on large, complex 3D regions, which exhibit obstacles (*e.g.*, buildings), varying terrain elevation, different coverage priorities, constraints on possible sensors placement. Our novel approach relies on computational geometry and statistical model checking, and enables the effective use of *off-the-shelf* AI-based black-box optimisers. Moreover, our method allows us to compute a *closed-form, analytical* representation of the region uncovered by a sensor deployment, which provides the means for *rigorous, formal certification* of the quality of the latter.**

**We show the practical feasibility of our approach by computing optimal sensor deployments for UAV localisation in two large, complex 3D critical regions, the Rome Leonardo Da Vinci International Airport (FCO) and the Vienna International Center (VIC), using NOMAD as our state-of-the-art underlying optimisation engine. Results show that we can compute optimal sensor deployments *within a few hours on a standard workstation and within minutes on a small parallel infrastructure*.**

## I. Introduction

The authors are with the Department of Computer Science, Sapienza University of Rome, 00198 Rome, Italy (e-mail: esposito@di.uniroma1.it; tmancini@di.uniroma1.it; tronci@di.uniroma1.it).

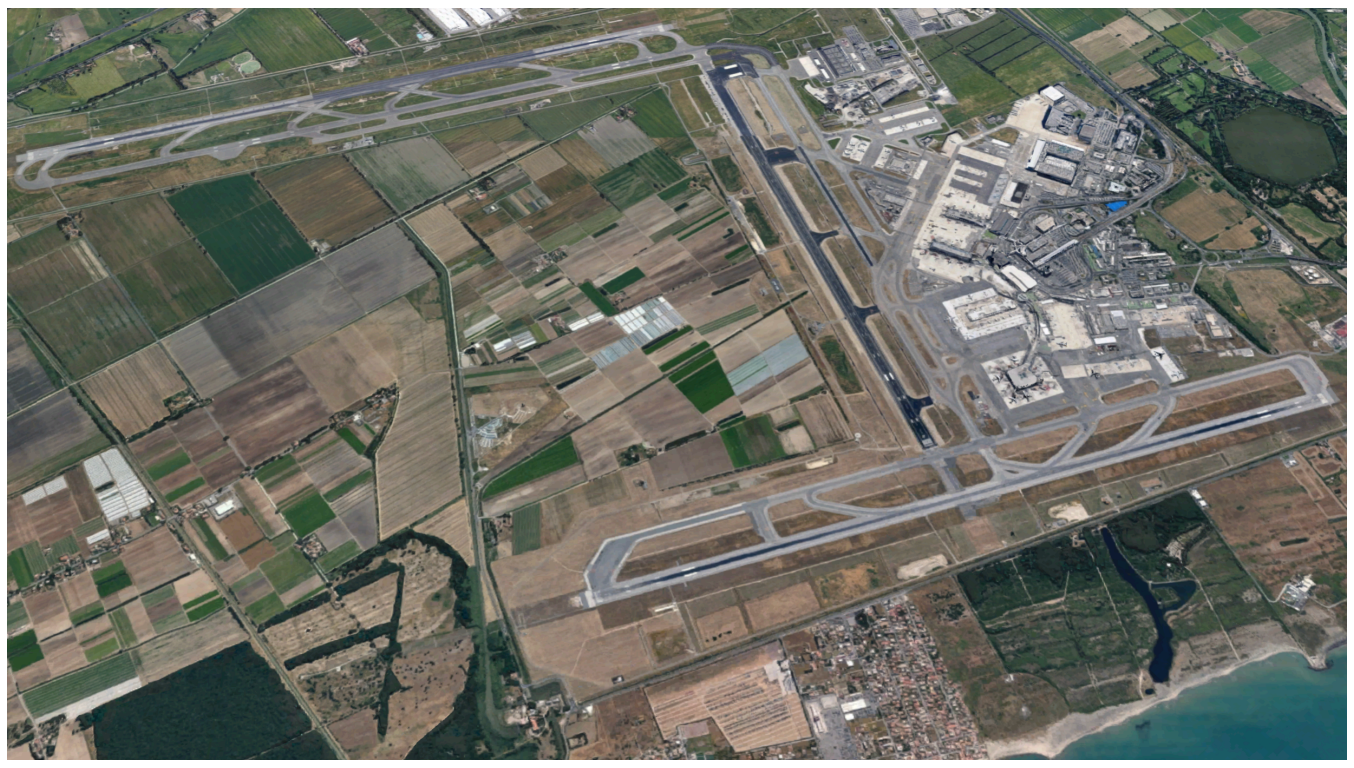

(a) Leonardo Da Vinci International Airport (FCO) of Rome

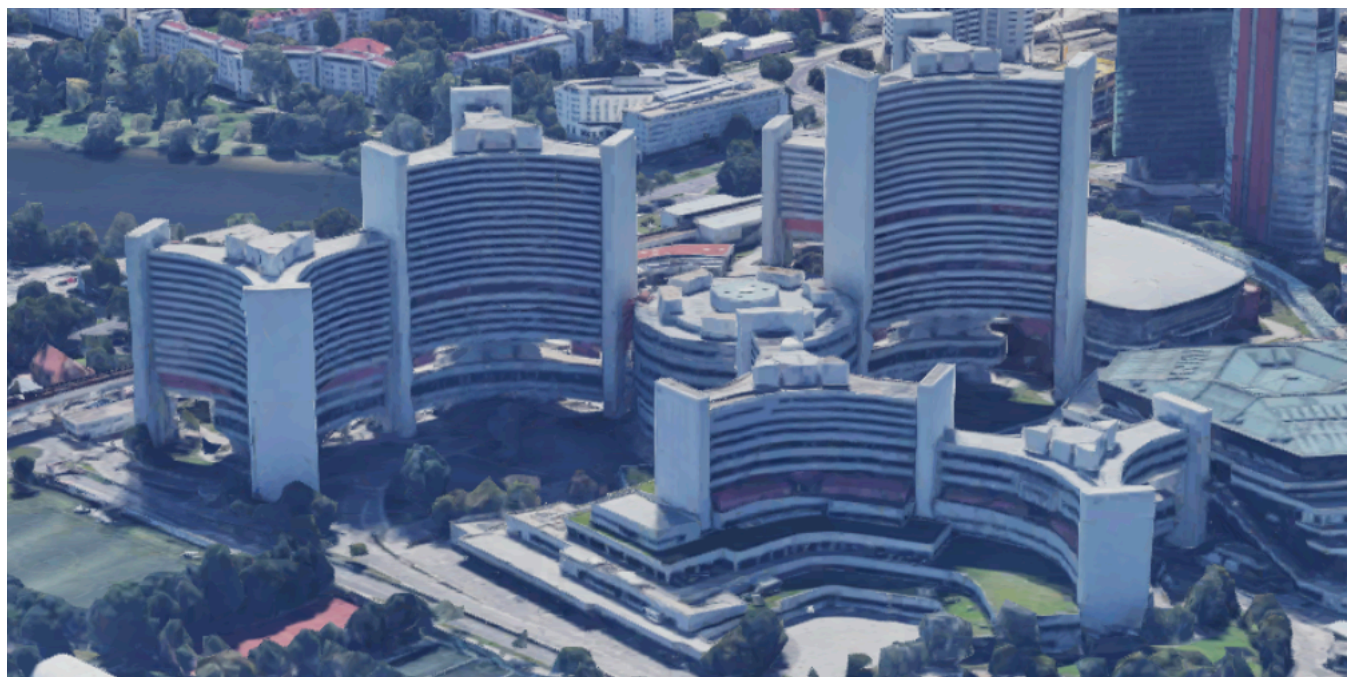

(b) Vienna International Center (VIC)

Figure 1: Satellite views of our case studies (Google ©).

In recent years, small Unmanned Aerial Vehicles (UAVs), or *drones*, have become widespread, given their increasingly lower prices and useful features. Besides the use of these devices for leisure (*e.g.*, photography) and public service activity like disaster monitoring, the public has become more and more aware of the risks that drones pose in several scenarios. A commercial drone, in fact, can disturb or jam radio communications, collide with other flying objects, perform espionage activity, and even carry offensive payloads like weapons or explosives. For these and other reasons, drones are seen as a *serious threat* in critical areas such as airports, military bases, correctional centres, power plants, and government sites. Figure 1 shows two examples of critical areas: the Rome Leonardo Da Vinci International Airport (FCO) in Italy and the Vienna International Center (VIC) in Austria, which we have used as our case studies.

Many surveillance solutions, based on a variety of technologies, have been designed to quantify and reduce the risks stemming from the presence of unauthorised drones in critical areas. In general, an anti-drone system has one or more goals, which may target both the drone and its controller: detection, localisation, identification, tracking, and countering. In this article, we focus on deployments of sensors aiming at *localising* unauthorised (possibly malicious) drones.

Each technology has its advantages and limitations. We refer the reader to [1], [2] for a survey. In our setting, *i.e.*, in the case of critical areas, an anti-drone system must have certain specific features, and this limits the suitability of most sensor typologies. Namely, the system must be able to detect drones over possibly large distances (*e.g.*, up to several kilometres within an airport); the localisation must be reliable with minimal to no false positives and false negatives; the system must be safely employable in public areas and at close distance to people. Such requirements rule out video and acoustic sensors, due to their low detection reliability, and radar systems for their high-power electromagnetic emissions, which may produce adverse health effects. Radio frequency sensors, which detect and localise drones based on the radio signals they emit, are among the most used technologies for dealing with the problem [1]. They operate over lower frequencies than radars and can effectively isolate the signals emitted by drones from other radio frequency signals originating from other sources (*e.g.*, WiFi). Direction-finding radio frequency sensors are able to localise drones via *triangulation* and offer high reliability over large distances (up to several kilometres in ideal conditions), with low to no impact from light and weather conditions.

### A. Motivation

The choice of the right sensor typology or of a mix of different typologies only solves a part of the general problem. When designing a system for drone localisation, it is crucial to deploy the sensors in the best possible way. The two main factors that influence the quality of an anti-drone system are its *cost-effectiveness* and its *ability to cover the Region of Interest (RoI)*. The latter is usually formalised as *coverage*, a measure of the extent of the portion of the monitored region where the deployed sensors can *effectively* localise targets. Typically, higher coverage requires higher costs, which stem from, *e.g.*, the need of a higher number of sensors or the necessity to deploy sensors in inconvenient positions (*e.g.*, on the roof or walls of a building, or where additional infrastructures must be built), which increase installation costs. We note that a single sensor, at the time of writing, may cost several tens of thousands of dollars.

The problem of *optimally* placing a set of sensors (with respect to a suitable blend of the conflicting objectives coverage and cost) has been widely explored in the literature, due to its relevance in many areas, *e.g.*, sensor networks, infrastructure security and safety, smart cities and smart homes. A discussion on related work is provided in Section II.

Determining a deployment of sensors that shows good coverage and cost-effectiveness measures may still not be enough for many critical scenarios. For example, more elaborate coverage measures may need to be considered, which take into account portions of the RoI with different *priorities*, multiple *sensing quality* levels, and tolerance to *sensor faults*. Also, in such settings, providing as output only aggregate – although quantitative– quality measures (*e.g.*, the ratio of the volume of the RoI satisfactorily covered), is often *not enough* to build trust in the quality of the deployment found, and delivering compelling evidence about *which* portions of the RoI are satisfactorily covered and which are not is crucial to provide means for rigorous auditing and formal certification of the quality of the deployment. It may be the case, for instance, that a deployment which appears to well cover a critical portion of the RoI is actually problematic, as it would not withstand the failure of a single sensor. Or that a deployment which covers most of the volume of the RoI still leaves drones the sufficient amount of freedom to move undetected via, *e.g.*, narrow, worm-shaped corridors (which, *e.g.*, could form only after the sabotage of a single sensor), and approach, still undetected, a critical target, thus causing serious damage. Note that, being able to determine coverage point by point, as in [3], [4], [5] in a discrete space is not enough to study this kind of properties of a deployment.

### B. Contribution

In this article, we present a novel approach to the sensor placement problem based on computational geometry and Artificial Intelligence (AI) Black-Box Optimisation (BBO).

In recent years, AI-based BBO techniques have witnessed exceptional improvements; new algorithms and tools are now able to tackle large and hard optimisation problems [6], [7], [8]. These optimisers make use of a black-box that implements the objective function and, possibly, the problem constraints. Thanks to powerful AI heuristics and surrogate models automatically learnt during search, these tools are powerful enough to optimise the objective, subject to the provided constraints, even if the black-box is computationally expensive to evaluate (our case).

We designed and developed Geometry-based Sensor Deployment Coverage Analyser (GD-Cover), a software tool that efficiently computes (via computational geometry and statistical model checking) quality metrics for a given sensor deployment, as well as a *closed-form, analytical* representation of the uncovered region which provides the means for *rigorous, formal certification* of its quality.

We show that, using GD-Cover as a black-box, the sensor deployment problem can be efficiently tackled by *off-the-shelf* AI-based BBO solvers (NOMAD [9] in our experiments). Our method scales very well over large and complex scenarios with many obstacles and over large numbers of sensors.

To the best of our knowledge, our approach is also the first to enable the computation of a closed-form 3D representation of the region not covered by a set of sensors. Finally, thanks to our proof-of-concept visualisation web app, such a representation can be navigated by the users, so to precisely understand the characteristics of deployments and to facilitate possible audits and certification.

## II. State of the art

The problem of determining the optimal placement of sensors inside a given region has been extensively studied in the literature. An *optimal placement* is defined as a positioning and configuration of a set of sensors that optimise given Key Performance Indicators (KPIs). Most studies consider the coverage performance metric, which measures the quality of a deployment based on the portion of the RoI that it can monitor. The configurations may vary depending on the kind of sensors; for instance, camera sensors can be configured by, *e.g.*, tweaking orientation, pan, tilt, and zoom. Due to the complexity of the problem and the large size of the scenarios of interest, exact optimisation approaches are not generally viable, and virtually all works in the literature exploit incomplete (best-effort) optimisation techniques, typically metaheuristic methods (*e.g.*, evolutionary algorithms and particle swarm optimisation) and various forms of gradient descent.

Sensor deployments are often very costly; therefore, many existing techniques also minimise the overall cost of the deployment. Other performance metrics have been studied, such as least exposure coverage, fault tolerance, and minimum overlapping (see, *e.g.*, [10], [11], [12], [13], [14], [5]).

Several works, *e.g.*, [15], [16], [17], [18], [19], [20], investigate the related problems of computing deployments of wireless sensor networks and of UAVs optimising additional KPIs such as connectivity, energy efficiency, and reliability.

Many existing studies assume a very simplified environmental setting, where the RoI is defined as a 2D space [21], [12], [13], [22], [23], [24], [25], even if, in some cases (*e.g.*, [26]), sensors can be deployed at different heights. These techniques cannot be applied in the case of UAVs localisation, where targets fly in a 3D space, as the coverage of a 2D region cannot be easily generalised to that of a large 3D scenario with obstacles.

Several approaches consider the problem of monitoring a 3D space, but with significant limitations affecting the applicability of these approaches to localising small UAVs in large critical areas, such as those considered here. In [5], [27], [28], [11], [14], [29], the candidate points for placement all lie on a 3D surface; in [30], [3], [4], [31], [32], [33], admissible sensor positions or points to be monitored (or both) belong to *finite* sets in the 3D space. In particular, in most existing 3D approaches, the RoI is discretised in cells. The visibility algorithms work on the assumption that if a point in the cell (*e.g.*, the centroid) is visible, then the whole cell is covered. This assumption becomes problematic in our setting. Consider, *e.g.*, our FCO case study: this environment has an area of around 16 km$^2$ and a height of 100 m. If we discretise the space into 3D cubic cells with edges of length 50 m, we would obtain around 12 800 cells (or target points). Although this number is still manageable by existing approaches, we note that a sensor deployment covering the centroid of a 125 000 m$^3$ cell does not guarantee that it can localise a 50 cm–long drone anywhere within the cell. Conversely, if we use much smaller cell edge lengths, such as 1 m, we would obtain approximately 1.6 *billion* cells. Such a large number would not be feasible to handle by any existing method with reasonable time and computational resources, even if exploiting graphical processing units, as in [34].

In practical applications, not all sensor deployments have the same economic cost. Typically, the cost depends not only on the number of deployed sensors, but also on their characteristics and positions. The minimisation of the number of sensors is studied in [4], [21], [22], [28], [31], [32]; however, it is often the case that the available sensors have different prices due to different characteristics, so minimising the number of sensors does not imply that the overall cost is minimised. Furthermore, the cost of physically deploying a sensor also depends on its position. For instance, mounting a heavy sensor on a wall costs more than mounting it on a roof, which in turn costs more than placing it on the ground. [3], [5], [35], [27] assume sensors have a fixed cost, whilst [36], [14] assign to each sensor a cost based only on the altitude and the roughness of the terrain in its position. Conversely, our approach handles this issue as a first-class citizen.

Summing up, to our knowledge (see also the recap in Appendix A), no other available approach optimises sensor deployments for UAV localisation in large, complex 3D regions with obstacles, varying terrain elevation, and in presence of constraints on admissible placements, by simultaneously taking into account sensors of different typologies, different placement costs, multiple sensing quality levels, fault tolerance. Also, no other approach supports the computation of a closed-form 3D representation of the region not covered by a candidate deployment.

## III. Problem modelling

In this section we present our geometric modelling approach to the computation of an optimal sensor deployment. In the following, $\mathbb{R}$, $\mathbb{R}_{0+}$, $\mathbb{R}_{+}$ denote the set of all, non-negative, and strictly positive real numbers, while $\mathbb{N}$ and $\mathbb{N}_{+}$ denote the set of non-negative and strictly positive integers.

Although our forthcoming definitions are well posed for real spaces of any number of dimensions, they will be given for regions of the 3D space $\mathbb{R}^3$, since this is what we need for our problem. We thus use the general term *region* to denote any set of points in $\mathbb{R}^3$. Also, given three points $A, B, C \in \mathbb{R}^3$, we define by $AB$ the straight-line segment between $A$ and $B$, by $\overline{AB}$ its length (*i.e.*, the Euclidean distance between $A$ and $B$), and by $\angle BAC \in [0, \pi]$ the angle formed by $AB$ and $AC$. Finally, given two regions $R, R' \subseteq \mathbb{R}^3$, the *distance between* $R$ *and* $R'$, notation *dist*$(R, R')$, is defined as $\min \left\{ \overline{XY} \mid X \in R, Y \in R' \right\}$. This notion naturally reduces to the Euclidean distance $\overline{XY}$ between two points $X$ and $Y$, if $R = \{X\}$ and $R' = \{Y\}$ are both singleton sets.

### A. Region of Interest (RoI) and obstacles

The RoI, *i.e.*, the region where the presence of unauthorised UAVs is to be detected, is some $\mathcal{R} \subset \mathbb{R}^3$ having *finite* volume.

The RoI can exhibit *obstacles*, *e.g.*, buildings, other artifacts, or simply varying ground elevations, which can hinder the radio visibility of some target points by a deployed sensor. Being able to explicitly model the position and shape of such obstacles is thus crucial to accurately evaluate the quality of a sensor deployment. We assume that the space occupied by obstacles in the RoI is defined as a (typically disconnected) region $\mathcal{O} \subset \mathcal{R}$.

### B. Priorities

For certain kinds of environments, the coverage of some portions of the RoI $\mathcal{R}$ is more important than that of others. In an airport, for instance, being able to localise a UAV flying above the runways could be more important than localising one close to the terminals.

We assume that $\mathcal{R}$ is *partitioned* into a finite number of regions having different priorities. Priorities are encoded as elements in finite set $\mathcal{H}$. The overall RoI $\mathcal{R}$ is hence partitioned into $\{\mathcal{R}_h \mid h \in \mathcal{H}\}$, where $\mathcal{R}_h$ is the portion of $\mathcal{R}$ having priority $h$. Being a partition, each point in $\mathcal{R}$ is assigned exactly one priority value.

### C. Sensor deployments, quality-guaranteed point coverage by triangulating sensors

The uncertainty in detecting a target provided by two triangulating sensors is known to vary with respect to the target-sensor distance and the angle $\theta$ between the target and the sensors (see, *e.g.*, [37]).

*Multiple sensing quality levels:* We support optimisation with respect to *multiple sensing quality levels*. To this end, we assume that a finite set $\mathcal{Q}$ is defined to denote different requested sensing quality levels. This set is *ordered* so that higher values in $\mathcal{Q}$ denote higher quality levels.

*Sensing angles:* For each $q \in \mathcal{Q}$, we assume that a *sensing angle range* $[\theta_{\min,q}, \theta_{\max,q}] \subset (0, \pi)$ is provided, defining *bounds* to be respected (Definition III.1) by the angle $\theta$ between the target and the two triangulating sensors, to ensure quality level $q$. For physical reasons, we can assume that $[\theta_{\min,q'}, \theta_{\max,q'}] \subseteq [\theta_{\min,q}, \theta_{\max,q}]$ for $q \leq q'$. Note that sensing angle ranges are within $(0, \pi)$ so that, to be localised with any useful accuracy, the target must not be *collinear* with the two sensors (as collinearity would hinder triangulation [37]).

*Sensors:* Each sensor $s$ is defined in terms of its *admissible positions* $\mathcal{A}_s \subseteq \mathbb{R}^3$ (*i.e.*, where it can be placed), the function $\text{cost}_s : \mathcal{A}_s \rightarrow \mathbb{R}_+$ defining the *cost to deploy the sensor* in each admissible position, and its *technical capabilities*. Admissible positions may not correspond with $\mathcal{R}$: indeed, there may be portions of $\mathcal{R}$ where sensors cannot be deployed (*e.g.*, bodies of water or airport runways) or, conversely, it may be possible to deploy sensors outside $\mathcal{R}$. The technical capabilities of $s$ are available in the form of a set $\{r_{s,q}, f_{s,q} \mid q \in \mathcal{Q}\}$. For each sensing quality level $q$, $r_{s,q} \in \mathbb{R}_+$ and $f_{s,q} \in \mathbb{R}_+$ are, respectively, the *maximum target distance* and the *radius of the First Fresnel Zone (FFZ)* needed by $s$ to detect a target UAV with the accuracy required to satisfy quality level $q$ (Definition III.1). The FFZ of sensor $s$ when detecting a target at position $X \in \mathcal{R}$ and ensuring quality level $q$ is the 3D ellipsoid whose main axis is the segment connecting the position of $s$ and $X$ and whose secondary axis has length $2f_{s,q}$. The FFZ radius of $s$ ($f_{s,q}$) is the greatest value such that, if a stray component of the signal transmitted by an UAV bounces off an object within the FFZ and then arrives at $s$, the resulting phase shift will be considered to have negative impact on the signal quality *incompatible* with sensing quality level $q$. Given the ranges on the transmitting/receiving power of the employed antennas and on the band used, as well the maximum distance $r_{s,q}$ for each $q \in \mathcal{Q}$, upper bounds to the values of $f_{s,q}$ for each $s$ and $q$ can be computed once and for all. Again, for physical reasons, we assume that $r_{s,q} \geq r_{s,q'}$ and $f_{s,q} \leq f_{s,q'}$ for $q \leq q'$.

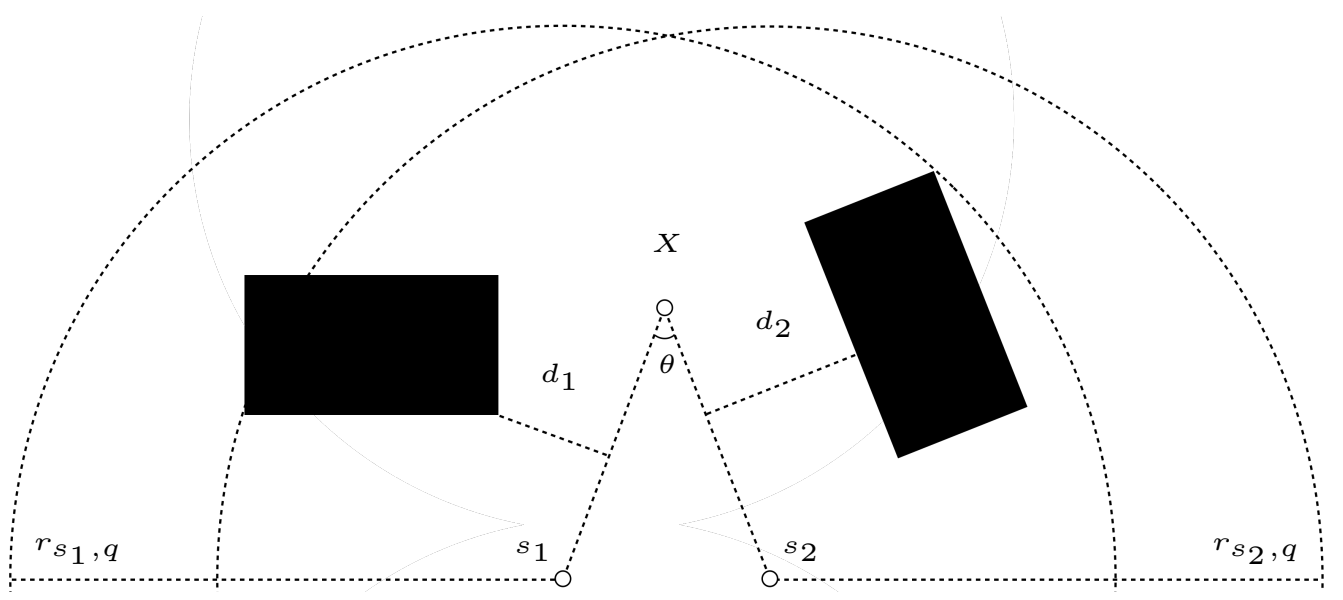

Figure 2: 2D $q$-coverage of point $X$ by two sensors, with two obstacles (in black); $d_i = \textit{dist}(X\mathcal{D}(s_i), \mathcal{O}) > f_{s_i,q}$, $i \in [1, 2]$.

*Sensor deployment:* We finally define a *deployment of a set of sensors* $\mathcal{S}$ a function $\mathcal{D}$ assigning an admissible position $\mathcal{D}(s) \in \mathcal{A}_s$ to each $s \in \mathcal{S}$.

Definition III.1 (see Figure 2 for an illustration in 2D) formalises our criterion to establish whether a point $X$ is covered by two sensors with quality at least $q$.

**Definition III.1** (Point $q$-coverage by triangulating sensors)**.** *Point $X \in \mathcal{R} - \mathcal{O}$ is* covered *by two triangulating sensors $s_1$ and $s_2$ of deployment $\mathcal{D}$* with quality at least $q$ *(in short: $X$ is $q$-covered by $s_1$ and $s_2$) if:*

*(1) $X$ is within the ranges of $s_1$ and $s_2$ for quality level $q$: $\textit{dist}(X, \mathcal{D}(s_i)) \leq r_{s_i,q}$, $i \in [1, 2]$;*
*(2) The line of sight between $X$ and each of the two sensors lies at a distance higher than the radius of the sensor's FFZ for $q$ from any existing obstacle: $\textit{dist}(X\mathcal{D}(s_i), \mathcal{O}) > f_{s_i,q}$, $i \in [1, 2]$;*
*(3) The angle $\theta$ formed by the points where the two sensors are placed and $X$ is within the sensing angle range for $q$: $\angle\mathcal{D}(s_1)X\mathcal{D}(s_2) \in [\theta_{min,q}, \theta_{max,q}]$.*

*We write* $\text{cover}^q(X, s_1, s_2) = 1$ *to denote that $X$ is $q$-covered by $s_1$ and $s_2$, and* $\text{cover}^q(X, s_1, s_2) = 0$ *otherwise.*

### D. Fault-tolerant quality-guaranteed coverage

In critical settings such as ours, *tolerance to sensor faults* is an important issue. Hence, we will define quality-guaranteed coverage even in (limited) presence of sensor faults. Namely, Definition III.2 defines the whole portion of the RoI *not* guaranteed to be covered by a sensor deployment $\mathcal{D}$ with a required quality level $q \in \mathcal{Q}$ when in presence of at most $j \geq 0$ (unknown) faulty sensors (in short: $(j, q)$-uncovered region).

**Definition III.2** ($(j, q)$-uncovered region)**.** *The set of points $X \in \mathcal{R} - \mathcal{O}$* $(j, q)$-uncovered *by a deployment $\mathcal{D}$ is:*

$$\mathcal{U}^{j,q} = \left\{ X \in \mathcal{R} - \mathcal{O} \;\middle|\; \textit{cover}^{j,q}(X) = 0 \right\} \quad (1)$$

*where:* $\textit{cover}^{j,q}(X) = \min_{\substack{F \in 2^{\mathcal{S}} \\ |F| \le j}} \max_{\substack{(s_1,s_2) \in (\mathcal{S}-F)^2 \\ s_1 \neq s_2}} \textit{cover}^{q}(X, s_1, s_2)$. *Namely,* $\textit{cover}^{j,q}(X)$ *is* $0$ *for those points* $X$ *for which there exists a set* $F$ *of at most* $j$ *sensors that, if faulty, would prevent* $q$*-coverage of* $X$ *by the others;* $\textit{cover}^{j,q}(X)$ *is* $1$ *otherwise.*

### E. Objective

Given a RoI $\mathcal{R}$ with obstacles $\mathcal{O}$, a partitioning of $\mathcal{R}$ into priority regions $\{\mathcal{R}_h \mid h \in \mathcal{H}\}$, sets $\mathcal{Q}$ (sensing quality levels) and $\mathcal{S}$ (sensors), and a maximum number $k > 0$ of sensors which can be faulty, our goal is to find a (admissible) deployment $\mathcal{D}^* = \mathcal{S} \rightarrow \mathbb{R}^3$ of $\mathcal{S}$ that minimises an objective function of the form:

$$\mathcal{D}^* = \arg\min_{\mathcal{D}} \; \textit{Placem}(\mathcal{D}) + \textit{Uncov}(\mathcal{D}) \quad (2)$$

given as the linear combination of the following (possibly conflicting) KPIs:

- $\textit{Placem}(\mathcal{D})$ is the sensors placement cost of $\mathcal{D}$, *i.e.*, the actual expense due to the chosen placement of the sensors:
$$\textit{Placem}(\mathcal{D}) = \sum_{s \in \mathcal{S}} \textit{cost}_s(\mathcal{D}(s)). \quad (3)$$
- $\textit{Uncov}(\mathcal{D})$ is the cost due to lack of coverage of $\mathcal{D}$, *i.e.*, the weighted volume of $\mathcal{R}$ not (satisfactorily) covered by $\mathcal{D}$, under at most $k$ faulty sensors:
$$\textit{Uncov}(\mathcal{D}) = \sum_{j=0}^{k} \sum_{q \in \mathcal{Q}} \sum_{h \in \mathcal{H}} w(j,q,h) \times \textit{volume}(\mathcal{U}^{j,q} \cap \mathcal{R}_h). \quad (4)$$

Each weight $w(j,q,h)$ denotes the implicit cost of not ensuring $(j,q)$-coverage of a unit of volume in $\mathcal{R}_h$.

Such two (possibly conflicting) KPIs are defined as to represent an amount of money, and this allows us to sum them up into a single objective value (2), the *Overall Deployment Cost (ODC)* of $\mathcal{D}$. Thus, the objective function would consider both the actual expense for the envisioned placement of the sensors in the chosen locations and the implicit costs due to their lack of (satisfactory) coverage.

## IV. Black-box optimisation

We cast the problem of finding an optimal sensor deployment as a *constrained optimisation* problem, where the *objective function* is (2), the *search space* is the set of assignments of 3D coordinates to each available sensor $s$, and where *constraints* enforce each sensor $s$ to be positioned within its admissible region $\mathcal{A}_s$ and to triangulate with at least one other sensor (*i.e.*, all sensors contribute to the coverage).

The complexity of computing the regions $\mathcal{U}^{j,q}$ (see (4)) needed to evaluate the objective function and their volumes hinders the possibility to exploit symbolic approaches (*e.g.*, Mixed Integer Linear Programming, Constraint Optimisation and the like) for scenarios of practical relevance, even those approaches explicitly aimed at solving very large instances, such as, *e.g.*, [38], [39].

We thus exploit AI-based BBO to solve realistic instances of the problem. BBO solvers make use of a *black box* that implements the objective function and the problem constraints. Thanks to powerful AI heuristics and surrogate models automatically learnt during search, these tools are powerful enough to optimise the objective, subject to the provided constraints, even if the black box is computationally expensive to evaluate (our case). In particular, state-of-the-art BBO solvers aim at *reducing* as much as possible the number of invocations of the (expensive) black box.

A BBO solver (we experimented with the state-of-the-art NOMAD [9] optimiser), repeatedly invokes our simulator (GD-Cover, see Section V) as a black box on multiple, intelligently-chosen candidate deployments. GD-Cover is in charge of computing both the objective value of the input deployment $\mathcal{D}$ and how much $\mathcal{D}$ violates the problem constraints (or, conversely, how robustly $\mathcal{D}$ satisfies such constraints). This combined feedback is then provided back to the BBO solver, which is in charge to find a better candidate deployment to submit to GD-Cover, also building and exploiting a local *surrogate model* of the solution space.

Given that our problem is highly non-linear and realistic instances are very large, finding a global optimum is an unviable option. BBO solvers like NOMAD are intrinsically incomplete, but guarantee *global convergence to local optima* and include sophisticated AI-based heuristics and *random restarts* to drive the search towards *high-quality optima*.

To exploit the availability of highly parallel computational infrastructures, possibly concurrently used for many tasks, we designed the two-level loosely-coupled parallel architecture shown in Figure 3. The overall system is assumed to work on overall $n$ computational nodes. An orchestrator process (left) samples a high number $N$ of random deployments (step 1) and asks GD-Cover (a highly-parallel process itself, see below) to evaluate their objective values (steps 2–3). These $N$ random deployments are then sorted from the best to the worst by the orchestrator (step 4) and then used (step 5) as initial assignments to launch, in parallel, $n_1 \le N$ BBO solvers, where random deployments are assigned to the available solvers in the order defined earlier (best initial deployments first), as soon as they become idle (this is equivalent to the sequential repeated invocation of a single optimiser using $N$ restarts, where the best random deployments are used first). Each of the $n_1$ solvers uses GD-Cover as its black-box. The best deployment found is given as the final solution. However, during the process, the current optimal deployment can be returned at any time, should the available time budget be over. GD-Cover itself (see Section V) is deployed as a highly-parallel computational service consisting of a front-end process for each of the $n_1$ BBO solvers, a centralised task dispatcher, and $n_2$ parallel helper processes, running on the remaining nodes (thus, $n_2 = n - n_1 - 2$).

## V. GD-Cover

In this section we present GD-Cover, a highly-parallel tool written in Java that, given a description of the geometry of the

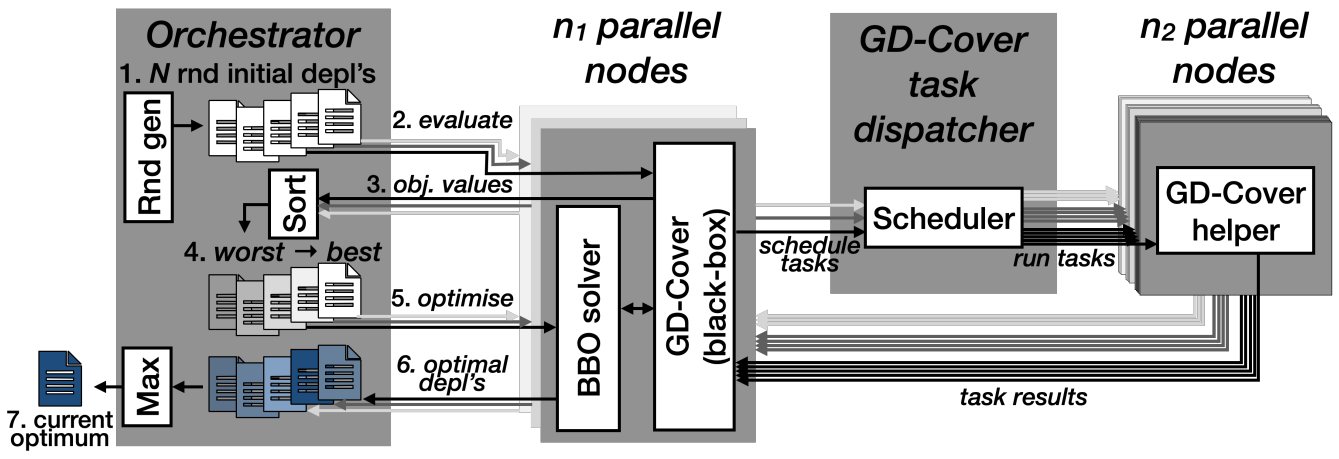

Figure 3: High-level architecture of our BBO-based approach.

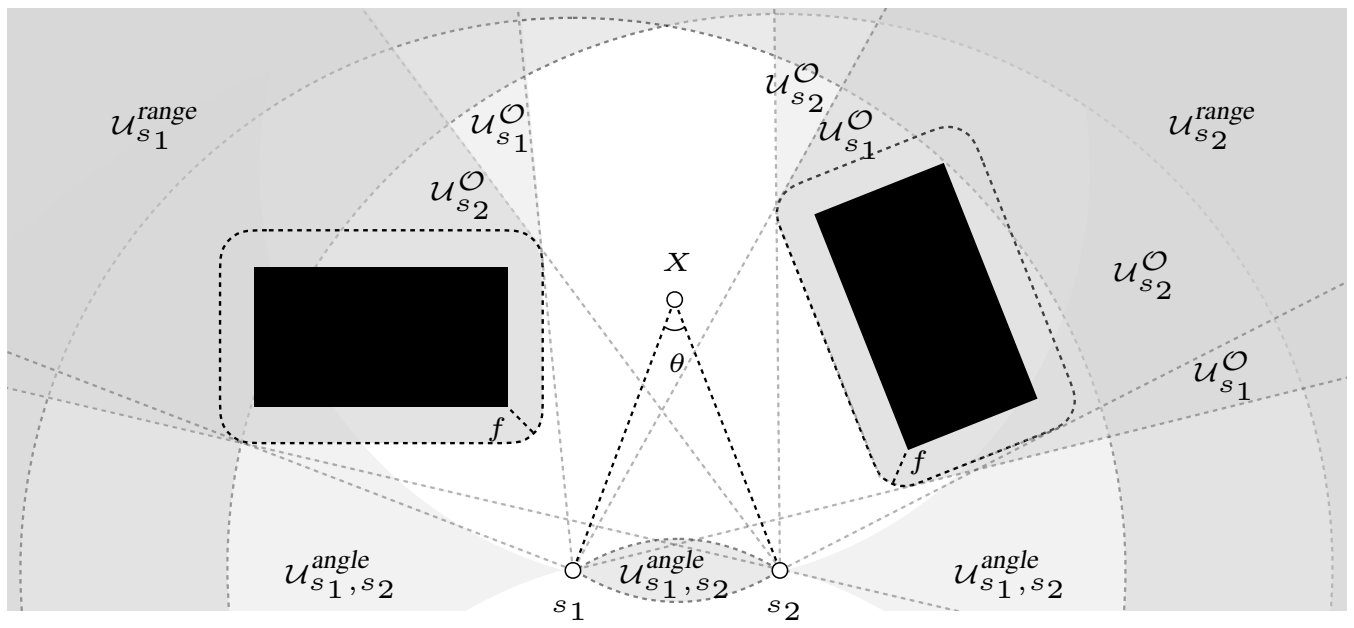

Figure 4: 2D example of region $\mathcal{U}_{s_1,s_2}$ uncovered by sensors $s_1$ and $s_2$ (grey region, formula (6)). Different elements of the union (6) are highlighted with different tones of grey. The figure assumes, for simplicity, $f_{s_1} = f_{s_2} = f$.

RoI $\mathcal{R}$ and of its obstacles $\mathcal{O}$ (Section III-A), a partitioning $\{\mathcal{R}_h \mid h \in \mathcal{H}\}$ of $\mathcal{R}$ in priority regions (Section III-B), a deployment $\mathcal{D}$ of a set of sensors $\mathcal{S}$ with known properties for a given set of sensing quality levels $\mathcal{Q}$, an angle interval $[\theta_{\min,q}, \theta_{\max,q}]$ for each $q \in \mathcal{Q}$, the maximum number $k$ of sensors that can be faulty, offers the following services (one or more services can be requested at the same time):

1. computes a quantitative measure of how much problem constraints are *violated* by $\mathcal{D}$ (thus proving that $\mathcal{D}$ is not admissible), or of how robustly they are *satisfied* (thus certifying that $\mathcal{D}$ is admissible);
2. computes a closed-form representation of the region $(j, q)$-uncovered by $\mathcal{D}$ for any $j \in [0, k]$ and $q \in \mathcal{Q}$;
3. efficiently estimates, by means of statistical model checking, the value of the objective function (Section III-E) with user-specified precision and statistical confidence.

GD-Cover has been designed to serve as a *black box* for our pool of BBO solvers seeking an optimal sensor deployment for a shared scenario. Hence, it is deployed as a highly distributed system, with each process being configured with a copy of the scenario of interest (*e.g.*, RoI, priorities, obstacles) and problem parameters. In particular, $n_1$ parallel processes are deployed, one per BBO solver, which act as front-ends (see Figure 3), and delegate the most intensive computations to a pool of $n_2$ helper processes, orchestrated by a centralised task dispatcher which guarantees adequate load balancing among the $n_1$ optimisation processes. The following sections describe the computations carried out by GD-Cover in more detail.

### A. Polyhedral geometry–based reasoning

GD-Cover performs its computations exploiting concepts from computational geometry. The key observation of the suitability of geometric reasoning to perform the computations above is forthcoming Proposition V.1, which shows how a representation of the $(j, q)$-uncovered region $\mathcal{U}^{j,q}$ (Definition III.2) can be provided in geometric terms, by relying on the following geometric notions: a) The *$\beta$-bloating of region $R$*, which is the set of points having distance at most $\beta \in \mathbb{R}_+$ from $R$: $bloat(R, \beta) = \{X \in \mathbb{R}^3 \mid dist(X, R) \leq \beta\} \supseteq R$; b) The *projection of point $X$ onto region $R$*, which is the region of points $Y$ such that the segment $XY$ intersects $R$: $proj(X, R) = \{Y \in \mathbb{R}^3 \mid XY \cap R \neq \emptyset\}$ (this is a specialised version of the projection defined in [40]). Note that, given point $X$ and region $R$, the set of points $Y$ such that the segment $XY$ has a *distance* from $R$ *at most* a given threshold $\beta$ can be defined as $proj(X, bloat(R, \beta))$.

**Proposition V.1** (Geometric representation of the $(j, q)$-uncovered region)**.** *The $(j, q)$-uncovered region of Definition III.2 can be equivalently defined as:*

$$\mathcal{U}^{j,q} = \bigcup_{\substack{F \in 2^{\mathcal{S}} \\ |F| \leq j}} \; \bigcap_{\substack{(s_1,s_2) \in (\mathcal{S}-F)^2 \\ s_1 \neq s_2}} \mathcal{U}^q_{s_1,s_2} - \mathcal{O} \tag{5}$$

*with*

$$\mathcal{U}^q_{s_1,s_2} = \left[\mathcal{U}^{\text{range},q}_{s_1} \cup \mathcal{U}^{\text{range},q}_{s_2}\right] \cup \left[\mathcal{U}^{\mathcal{O},q}_{s_1} \cup \mathcal{U}^{\mathcal{O},q}_{s_2}\right] \cup \mathcal{U}^{\text{angle},q}_{s_1,s_2} \tag{6}$$

*where, for $i \in [1, 2]$:*

- $\mathcal{U}^{\text{range},q}_{s_i}$ *is the* region out of range *of $s_i$:*
  $$\mathcal{U}^{\text{range},q}_{s_i} = \{X \in \mathcal{R} \mid dist(X, \mathcal{D}(s_i)) > r_{s_i,q}\},$$
  *i.e., the complement of a sphere of radius $r_{s_i,q}$ centred in the position of sensor $s_i$, $\mathcal{D}(s_i)$.*
- $\mathcal{U}^{\mathcal{O},q}_{s_i}$ *is the region* not covered *by $s_i$* because of obstacles. *It is the set of points $X \in \mathcal{R}$ such that the distance between an obstacle (a point in $\mathcal{O}$) and the straight-line segment connecting $X$ with $\mathcal{D}(s_i)$ is at most the FFZ radius of $s_i$ for quality level $q$, $f_{s_i,q}$:*
  $$\mathcal{U}^{\mathcal{O},q}_{s_i} = proj(\mathcal{D}(s_i), bloat(\mathcal{O}, f_{s_i,q})) \cap \mathcal{R}.$$
- $\mathcal{U}^{\text{angle},q}_{s_1,s_2}$ *is the* region not covered *by $s_1$ and $s_2$* because of excessive sensing error. *It is the set of points $X \in \mathcal{R}$ such that the angle $\angle\mathcal{D}(s_1)X\mathcal{D}(s_2)$ formed by $X$ with the positions of the two sensors is outside range $[\theta_{min,q}, \theta_{max,q}]$:*
  $$\mathcal{U}^{\text{angle},q}_{s_1,s_2} = \{X \in \mathcal{R} \mid \angle\mathcal{D}(s_1)X\mathcal{D}(s_2) \notin [\theta_{min,q}, \theta_{max,q}]\}.$$

Proofs are delayed to Appendix B. Figure 4 illustrates Proposition V.1 in 2D and under no faults (*i.e.*, $j = 0$), by showing the region uncovered by the sensors in Figure 2.

To carry out its tasks efficiently, GD-Cover approximates the environment, *i.e.*, the RoI and obstacles thereof, and performs all its computations in terms of bounded and unbounded *convex polyhedra* in $\mathbb{R}^3$. Polyhedral representations are indeed standard when handling data about geographic areas, terrain asperities, buildings, and other kinds of artifacts and shapes (*e.g.*, the geometry of the runways at an airport) within Geographic Database Systems and Computer Aided Design tools. Such a representation has several advantages: (i) every 3D region can be (both over- and under-) approximated with arbitrary precision by a set of convex polyhedra (typically a small number of polyhedra guarantees good approximations); (ii) since convex polyhedra can be defined via linear constraints, they are easy to manipulate efficiently using standard computational geometry techniques and libraries.

A union of convex polyhedra, in general, is not a convex polyhedron. However, there are well-known techniques to manipulate unions (*i.e.*, sets) of convex polyhedra very efficiently (some are briefly outlined in Section V-C).

Being able to represent arbitrary regions with unions of polyhedra yields a very convenient framework to perform complex operations. In fact, the union or the intersection of unions of polyhedra can still be efficiently computed as a union of polyhedra. In the case of the difference and complement operations, the result is a union of non-closed polyhedra; however, for our purposes such regions can be safely over- or under-approximated with unions of closed polyhedra with arbitrary precision. In the sequel, we will refer to convex closed polyhedra simply as *polyhedra*, and use the term *polyhedral representation* of a region to signify that the region is defined as a *union of closed convex polyhedra*.

Although the GD-Cover primary inputs are polyhedral, some of the computed regions (mainly those described in Proposition V.1) are non-polyhedral. GD-Cover computes polyhedral (under- and over-) approximations for them accurate up to a user-specified *error threshold* $\rho \in \mathbb{R}_+$. This will be the *maximum Euclidean distance* between a non-polyhedral region (*e.g.*, $\mathcal{U}_s^{\text{range},q}$, $\textit{bloat}(\mathcal{O}, f_{s,q})$, and $\mathcal{U}_{s_1,s_2}^{\text{angle},q}$ as defined in Proposition V.1, with $s, s_1, s_2 \in \mathcal{S}, q \in \mathcal{Q}$) and its computed polyhedral (under- and over-) approximations.

### B. Evaluation of constraints

The sensor positioning requirements outlined in Section IV break down to a number of problem constraints. GD-Cover evaluates each of them to a positive value when violated (in which case the resulting value is an indication of how much the constraint is violated), and to a zero-or-negative value when satisfied (in which case the resulting value is an indication of how robustly the constraint is satisfied with respect to perturbations of the candidate deployment).

Let $q_0$ be the *lowest* sensing quality level in $\mathcal{Q}$. For each sensor $s \in \mathcal{S}$, constraints are as follows:

*1) Sensor is placed not within or too close to obstacles:* If $\mathcal{D}(s) \in \textit{bloat}(\mathcal{O}, f_{s,q_0})$ (*i.e.*, if $s$ is within an obstacle or too close to an obstacle even for the lowest sensing quality level of interest), then the constraint is declared *violated* with cost $\textit{dist}(\mathcal{D}(s), \mathcal{R} - \textit{bloat}(\mathcal{O}, f_{s,q_0}))$. Otherwise, the constraint is declared *satisfied* with robustness value: $-\textit{dist}(\mathcal{D}(s), \textit{bloat}(\mathcal{O}, f_{s,q_0}))$.

*2) Sensor is placed within its admissible region:* If $\mathcal{D}(s) \notin \mathcal{A}_s$ (*i.e.*, if $s$ is positioned outside its admissibility region), then the constraint is declared *violated* with cost $\textit{dist}(\mathcal{D}(s), \mathcal{A}_s)$. Otherwise, the constraint is declared *satisfied* with robustness value: $-\textit{dist}(\mathcal{D}(s), \mathcal{R} - \mathcal{A}_s)$.

*3) Sensor is not isolated:* The constraint evaluates to $d = \min_{s' \in \mathcal{S} - \{s\}} \big(\textit{dist}(\mathcal{D}(s), \mathcal{D}(s')) - r_{s,q_0} - r_{s',q_0}\big)$. Thus, if $d > 0$, then the constraint is declared *violated* with cost $d$, which is an indication of how much $s$ must be moved to become non-isolated. Otherwise, if $d \leq 0$, then the constraint is declared *satisfied* with robustness value $d$, which is an indication of how much $s$ should be moved to become too far with respect to all sensors with which $s$ could now triangulate.

### C. Computing closed-form polyhedral approximations of the uncovered region

Here we show how GD-Cover computes polyhedral representations of $\mathcal{U}^{j,q}$, the regions $(j,q)$-uncovered (for $j \in [0,k]$ and $q \in \mathcal{Q}$) by the specific deployment $\mathcal{D}$ of sensors $\mathcal{S}$ given as input, as defined in Proposition V.1. From this representation, it will be easy to compute any kind of additional quality metrics of the input sensor deployment, hence also *any* (computable) objective function. This makes our approach extremely flexible (but see Section V-D).

Proposition V.1 defines region $\mathcal{U}^{j,q}$ as unions of intersections of a number of regions, $\mathcal{U}^q_{s_1,s_2}$, one for each pair of distinct sensors $s_1$ and $s_2$, from which the region occupied by the obstacles ($\mathcal{O}$) must be removed. Each $\mathcal{U}^q_{s_1,s_2}$ in turn is defined by a union of 5 regions: $\mathcal{U}_{s_1}^{\text{range}}$, $\mathcal{U}_{s_2}^{\text{range},q}$, $\mathcal{U}_{s_1}^{\mathcal{O},q}$, $\mathcal{U}_{s_2}^{\mathcal{O},q}$, $\mathcal{U}_{s_1,s_2}^{\text{angle},q}$, which are *not* polyhedral (see Proposition V.1). GD-Cover thus computes *polyhedral approximations* for them. To overcome the possible errors in such approximations, the tool can compute both polyhedral under- and over-approximations of such regions, using the value $\rho$ given as input (see Section V-A) as tolerance. Such under- and over-approximations in turn allow the tool to compute both a polyhedral under-approximation $\lfloor \mathcal{U}^{j,q} \rfloor$ and a polyhedral over-approximation $\lceil \mathcal{U}^{j,q} \rceil$ of the entire uncovered region $\mathcal{U}^{j,q}$. Hence: $\lfloor \mathcal{U}^{j,q} \rfloor \subseteq \mathcal{U}^{j,q} \subseteq \lceil \mathcal{U}^{j,q} \rceil$.

Thus, points in $\mathcal{R}$ belonging to $\lfloor \mathcal{U}^{j,q} \rfloor$ are *certainly* $(j,q)$-*uncovered* by the given sensor deployment, points outside $\lceil \mathcal{U}^{j,q} \rceil$ are *certainly* $(j,q)$-*covered*, while points lying in $\lceil \mathcal{U}^{j,q} \rceil - \lfloor \mathcal{U}^{j,q} \rfloor$ are *possibly* $(j,q)$-*uncovered*, with the uncertainty due to the possible errors introduced when computing polyhedral approximations of $\mathcal{U}_{s_1}^{\text{range},q}$, $\mathcal{U}_{s_2}^{\text{range},q}$, $\mathcal{U}_{s_1}^{\mathcal{O},q}$, $\mathcal{U}_{s_2}^{\mathcal{O},q}$, $\mathcal{U}_{s_1,s_2}^{\text{angle},q}$ for all pairs of distinct sensors $s_1$ and $s_2$. Priorities of $\mathcal{R}$ ($\{\mathcal{R}_h \mid h \in \mathcal{H}\}$, Section III-B) can be straightforwardly considered on top of $\lfloor \mathcal{U}^{j,q} \rfloor$ and $\lceil \mathcal{U}^{j,q} \rceil$: the uncovered portion of the RoI with priority $h$ is sandwiched between $\lfloor \mathcal{U}^{j,q} \rfloor \cap \mathcal{R}_h$ and $\lceil \mathcal{U}^{j,q} \rceil \cap \mathcal{R}_h$.

GD-Cover exploits the C++ Parma Polyhedra Library [41] for the efficient manipulation of convex polyhedra. Unfortunately, most of the needed computations suffer from *combinatorial explosion* in the worst case. To this end, GD-Cover takes clever countermeasures to handle realistic scenarios efficiently (see Appendix C for details). For example, when performing operations on intersections of unions of polyhedra (which could need to consider all tuples of polyhedra, one per union being considered, and could result in an algorithm whose time complexity in the worst-case is exponential in the number of such polyhedra) and to efficiently seek polyhedra of interest for the various computations, GD-Cover implements a *spatial indexing method* based on AABB Trees [42], which greatly reduces the number of operations needed in most cases (see Algorithm 1 in Appendix C).

To keep the size of the unions of polyhedra manipulated by the algorithm small, and to mitigate the quadratic explosion arising when considering all possible pairs of sensors, GD-Cover performs parallel computation exploiting helper processes via a centralised task dispatcher aimed at keeping load balancing (see Figure 3 and Algorithm 2 in Appendix C). Also, GD-Cover partitions the RoI in $m$ identical cells ($\mathcal{R}_1, \ldots, \mathcal{R}_m$) and delegates again the available helper processes to compute the $(j, q)$-uncovered portion of each $\mathcal{R}_c$ ($c \in [1, m]$), $\mathcal{U}_c^{j,q}$. Indeed, computing each single $\mathcal{U}_c^{j,q}$ is way faster than computing the entire $\mathcal{U}^{j,q}$, because, on average, the sizes of the unions of polyhedra manipulated by algorithm are smaller and the distances involved (much higher than the range of each sensor) imply that several sensors (and sensor *pairs*) are too far to possibly contribute to the coverage of the considered cell and can be excluded upfront. As a consequence, the computation of several of the $\mathcal{U}_c^{j,q}$s, also thanks to the AABB Tree-based spatial indexing, takes negligible time. Dynamic load balancing is dealt with by taking $m$ much larger than the number of helper processes, an approach which is trivial to implement (see, *e.g.*, [43], [44], [10]). Section VI-C2 experimentally evaluates the scalability of this parallelisation technique for the problem at hand.

We also implemented a proof-of-concept web app which, given the output of GD-Cover (*e.g.*, the uncovered region for the optimal deployment computed by the BBO solver), allows the user to visually and interactively navigate the RoI as a 3D space, see where sensors are actually planned to be deployed and which portions of the RoI are (un)covered, together with their priorities (details are delayed to Appendix E). All this enables any interactive analyses of the results, *e.g.*, visually inspecting any dangerous uncovered regions, *e.g.*, worm-shaped corridors which could be used by an attacker to move across the RoI undetected.

### *D. Statistical model checking–based estimation of the objective value*

By computing closed-form polyhedral representations of the uncovered regions $\mathcal{U}^{j,q}$, any objective function can be evaluated by analysing such regions. However, this computation is an intensive task (see Section VI-C2) and is not strictly needed to effectively guide the optimisation process, when only the objective value and an evaluation of the problem constraints is needed by the BBO solvers.

GD-Cover uses statistical model checking techniques (see, *e.g.*, [45] for a survey) to estimate the value of the objective function (2) efficiently via Monte Carlo sampling, while offering *statistical guarantees* on the accuracy of the approximation. Note that sampling points on a 3D fixed-step grid, instead that in the whole (continuous) RoI, would not yield any guarantees on the accuracy of the estimation beyond the grid step length, and would not be as effective in guiding the optimisation. In Section VI-C1 we show that the approximation of the objective value requires only a tiny fraction ($\ll 1\%$ in our case studies) of the time required to compute the uncovered regions in closed form, and so the objective function exactly. Hence, GD-Cover computes the uncovered regions in closed form only for the final (optimal) deployment and upon explicit user request.

To compute such an approximation of the objective value, GD-Cover uses a Monte Carlo–based algorithm along the lines of [46]. Namely, it combines the EBGStop approximation algorithm [47] and the hypothesis testing technique from [48]. Given values for two parameters, $\varepsilon, \delta \in (0, 1)$, the algorithm computes an $(\varepsilon, \delta)$-approximation of the mean value $\mu$ of a bounded random variable $Z$ *i.e.*, a value $\hat{\mu}$ guaranteed to lie within $\mu(1 \mp \varepsilon)$ with probability at least $(1 - \delta)$. The algorithm iteratively generates (again exploiting the available pool of parallel helpers, via the intercession of the task dispatcher to handle load balancing among the $n_1$ optimisation processes) i.i.d. samples of $Z$ until the termination condition of [46] is satisfied, which implies that the objective value estimated from the samples is a $(\varepsilon, \delta)$-approximation of the true value.

This approach can be used whenever the objective value for a candidate deployment can be expressed as the expected value of a bounded random variable $Z$. Definition V.1 and Observation V.1 show that this is the case for the objective function in (2). The complexity of (2) (which considers multiple sensing quality levels as well as fault tolerance) also indirectly shows that such an approach is very flexible, and many other objective functions fall in this class (if not, our BBO-based approach can still be used, but GD-Cover must be asked to compute the uncovered regions in polyhedral form on each candidate deployment to enable the computation of the objective values, leading to longer optimisation times).

**Definition V.1.** *Given $\mathcal{R}$, $\mathcal{D}$, $\mathcal{H}$, $\mathcal{Q}$, $\mathcal{U}$, and $\mathcal{R}_h$ ($h \in \mathcal{H}$) as in Section III-E, let $V_\mathcal{R}$ = volume($\mathcal{R}$) (a constant).*

*Let also $(\Omega^{j,q}, \mathcal{F}^{j,q}, Pr^{j,q})$ ($j \in [0, k], q \in \mathcal{Q}$) be the probability spaces such that:*

- $\Omega^{j,q} = \{\perp\} \cup \{h \mid h \in \mathcal{H}\}$ *is the space of outcomes* ($\perp \notin \mathcal{H}$)
- $\mathcal{F}^{j,q} = 2^{\Omega^{j,q}}$ *is the space of events*
- $Pr^{j,q} : \mathcal{F}^{j,q} \to [0, 1]$ *is the following probability measure:*
  – $Pr^{j,q}(\perp) = 1 - \frac{\text{volume}(\mathcal{U}^{j,q})}{V_\mathcal{R}}$
  – $Pr^{j,q}(h) = \frac{\text{volume}(\mathcal{U}^{j,q} \cap \mathcal{R}_h)}{V_\mathcal{R}}$ *for* $h \in \mathcal{H}$
  – $Pr^{j,q}(E) = \sum_{\omega \in E} Pr^{j,q}(\omega)$ *for any* $E \subseteq \Omega^{j,q}$.

Since $\{\mathcal{R}_h \mid h \in \mathcal{H}\}$ is a partition of $\mathcal{R}$, $Pr^{j,q}(\Omega^{j,q}) = 1$.

**Observation V.1.** *For every $j \in [0, k], q \in \mathcal{Q}$, let $Z^{j,q}$ be a real-valued random variable defined on probability space $(\Omega^{j,q}, \mathcal{F}^{j,q}, Pr^{j,q})$ (Definition V.1) as: $Z^{j,q}(\perp) = 0$; $Z^{j,q}(h) = V_\mathcal{R} \times w(j, q, h)$ (for $h \in \mathcal{H}$).*

*The value of the objective function (2) evaluated for deployment $\mathcal{D}$ is Placem$(\mathcal{D}) + \sum_{j=0}^{k} \sum_{q \in \mathcal{Q}} \mathbb{E}(Z^{j,q})$, where $\mathbb{E}^{j,q}(Z)$ is the* expected value *of $Z^{j,q}$.*

Observation V.1 is proved in Appendix B. Random variables $Z^{j,q}$, being bounded, clearly meet the requirements for the application of the statistical model checking algorithm described above. To generate i.i.d. samples for $Z^{j,q}$, each delegated helper (running in parallel) samples points $X \in \mathcal{R}$ *uniformly at random*. For each sample $X$, the helper determines whether $X$ is $(j, q)$-covered for every $j$ and $q$ (*or* falls within an obstacle, notation *cover*$^{j,q}(X) = 1$) or not (*cover*$^{j,q}(X) = 0$) by deployment $\mathcal{D}$. This is implemented by exploiting standard polyhedral geometry operations (note that all conditions of Definitions III.1 and III.2 can be checked in this way, simply by looping through the set of pairs of sensors). Value for each random variable $Z^{j,q}$ is computed from $X$ as follows: $Z^{j,q} = V_{\mathcal{R}} \times w(j,q,h) \times (1 - \mathit{cover}^{j,q}(X))$, where $h$ is the (single) priority value of point $X$.

## VI. Experiments

We exercised our parallel system by computing optimal anti-drone sensor deployments on two real-world case studies, the Leonardo Da Vinci International Airport (FCO) in Rome, Italy, and the Vienna International Center (VIC) in Vienna, Austria, described below, having complementary properties: whilst the former consists in a large environment with wide open spaces and relatively low obstacles, the latter presents several tall buildings condensed in a small area. In all experiments we used NOMAD v. 3.9.1 as our BBO solver.

### A. Experimental setting

*1) Sensors:* For each case study, we considered sensors of two types (named T1 and T2), with different characteristics (chosen in accordance to reference values from the literature, *e.g.*, [1]), and costs. For generality, we normalised all costs for each case study to the price of a single sensor of type T1 (the cheapest type) simply installed on a pole on the ground, at a height between 5–10 m (the cheapest installation). We sought for optimal fault tolerant quality-guaranteed coverage with two sensing quality levels ($\mathcal{Q} = \{q_0, q_1\}$, with $q_0 < q_1$) and one possible sensor fault (*i.e.*, $k = 1$).

*2) Priority regions:* The RoI of each case study was partitioned into two priority regions (low and high priority).

*3) Weights of the objective function:* Values of $w(j,q,h)$ (the cost of not $(j,q)$-covering a unit of volume of the region having priority $h$, Section III-E) are reported in Appendix D-A2. For example, in FCO, weights model *indifference criteria* such as: an increase of one volume unit of the $(0,q_1)$-covered high priority region ($w(0,q_1,\text{high}) = 20$) is equally exchanged with an increase of two volume units of the $(0,q_0)$-covered low priority region ($w(0,q_0,\text{low}) = 10$).

*4) Approximation thresholds and initial deployments:* The thresholds $\varepsilon$ and $\delta$ used by GD-Cover to estimate the objective value for each candidate deployment produced during optimisation were both set to 1%, thus guaranteeing that, with statistical confidence 99%, the estimated objective value is within a 1% error margin from its true value. Threshold $\rho$ used to compute polyhedral under- and over-approximations of the regions mentioned in Section V-A was set to 10. Finally, the number of random deployments generated and fed as starting points to the multiple BBO solvers was set to $N = 100$.

*5) Computational infrastructure:* Experiments were performed on a cluster of identical machines, each one equipped with 2 AMD EPYC 7301 CPUs (overall 64 cores) and 256GB RAM. Our loosely-coupled architecture (Figure 3) is particularly suited for off-premise clusters as ours, which are shared among a high number of competing processes (a common paradigm aimed at keeping the cost of running parallel software low). Below, we present the completion times that would be obtained in three reference infrastructures, namely *if fully reserving* the following number of machines: (i) 1 ($n = 64$, using $n_1 = 5$ BBO solvers and the remaining $n_2 = 57$ nodes as GD-Cover helpers); (ii) 10 ($n = 640$, $n_1 = 20$, $n_2 = 618$); (iii) 50 ($n = 3200$, $n_1 = 100$, $n_2 = 3098$). A full scalability analysis is delayed to Appendix D-B.

### B. Case studies

*1) Leonardo Da Vinci International Airport (FCO):* This is a prototypical example of a critical area, with a total surface of approximately 16 km$^2$, most of which taken by the three 4 km-long runways. We created a simplified 3D model of the RoI as the union of 11 polyhedra with a height of 100 m. The total volume to be monitored is thus 1.6 km$^3$. The case study presents many obstacles, modelled with 52 polyhedra, ranging from large buildings such as hangars and terminals to small service buildings. Figure 1a shows an aerial view of FCO, whilst Figure 6 shows screenshots of our visualiser with our 3D model.

Sensor costs are 1 (for type T1, reference cost) and 1.5 (T2). Sensors can be placed everywhere on the ground (except for the runways), and over the walls and roofs of (most of) the obstacles (in the latter cases with cost overheads of 10% and 20%, respectively). Sensors of type T1 (respectively, T2) can detect targets distant up to 1000 m (respectively, 1250 m) with quality level $q_0$, and up to 900 m (respectively, 1110 m) with quality level $q_1$. Sensing angle ranges for any sensor pair are $[25°, 155°]$ (for $q_0$) and $[30°, 150°]$ (for $q_1$). All sensors have the same FFZ (5 m) for both $q_0$ and $q_1$. The high-priority portion of the RoI includes the runways and the space above them, while the remaining region is low-priority.

*2) Vienna International Center (VIC):* This is a complex of several buildings that hosts the headquarters of important organisations of the United Nations. Its political and economic relevance make this environment a critical area. VIC has a substantially different structure than FCO: while the total volume is only a fraction of the volume of FCO, the RoI is much more densely occupied by tall buildings. This makes the optimal placement problem harder, since each sensor will only cover a small portion of the region, regardless of its position, due to lack of line-of-sight visibility. Our 3D RoI model is a cube with edges of length 400 m containing simplified shapes for all the relevant buildings (modelled with 51 polyhedra).

Sensor costs are 1 (type T1, reference cost) and 1.17 (T2). All sensors can be placed on the ground, and T2 sensors also over the roofs and on the concrete walls of the towers, *i.e.*, not on the glass façades (in the latter cases with cost overheads of 5% and 10%, respectively). Sensors of type T1 (respectively, T2) can detect targets up to 500 m (respectively, 700 m) with quality level $q_0$, and up to 400 m (respectively, 600 m) with quality level $q_1$. Sensing angle ranges for any sensor pair are $[25°, 155°]$ (for $q_0$) and $[30°, 150°]$ (for $q_1$). All sensors have the same FFZ (5 m) for both $q_0$ and $q_1$.

The high-priority portion of the RoI is composed of two parts: the first one is a portion of a spherical shell on top of the buildings, constituting a sort of dome on the RoI; such region is deemed as highly important because, in such a small environment (densely occupied by buildings), it is crucial to detect the arrival of an UAV as quickly as possible. The second portion includes the area at ground-level, that is the busiest area where people walk to enter the buildings.

Figure 1b shows an aerial view of VIC, whilst Figure 6 shows screenshots of our visualiser with our 3D model.

### C. Experimental results

*1) Optimisation:* We ran multiple experiments for each case study, and with various *configurations*, where each configuration defines an overall number of sensors and the relative numbers of T1 vs. T2 sensors. We used between 10 and 20 sensors for FCO and between 3 and 10 for VIC. Indeed, preliminary experiments shown that higher numbers of sensors would not bring any significant further improvement of the coverage, while fewer sensors would simply be ineffective.

Figure 5 (up) shows the Overall Deployment Costs (ODCs) of the best deployments found for each configuration (T1 vs. T2 sensors). The darker a cell in the heat-maps, the better the quality of the final deployment found for that configuration. Numbers in cells denote the objective value (2) of the optimal deployments found. Values in parenthesis denote the best solution for each number of sensors. The value with a "*" (in the darkest cell) denotes the best solution overall.

Our system achieved a final (optimal) deployment yielding, on average across the various configurations, an expected reduction of the ODC of 27.37% (FCO) and 26.69% (VIC), where the expectation is computed with respect to the $N$ initial random generated admissible deployments (time zero). Expected ODC reductions on the best configurations are of 31.89% (FCO) and 32.01% (VIC).

Figure 5 (down) shows, for each configuration and each timepoint $t$, the objective value of the best deployment that our system (when running on infrastructure (ii), *i.e.*, on 640 *fully reserved* for the job) would have found if halted at time $t$. The bold curves refer to the configurations of T1 vs. T2 sensors yielding the overall optimum, namely 13 T1 and 3 T2 sensors for FCO, and 4 T1 and 3 T2 sensors for VIC (see Figure 5, up). The objective values for each configuration have been normalised as percentages of the objective value of the final deployment found for that configuration. The system terminated in 11m45s (FCO) and 18m18s (VIC), and found a solution whose objective value is just 10% above the final optimum in only 34s (FCO) and 6s (VIC). An analysis of the scalability of our parallel system on fully reserved infrastructures of various sizes is delayed to Appendix D-B. Here we just mention that, on infrastructure (iii) (*i.e.*, using 3200), it would have terminated in only 3m5s (FCO) and 3m40s (VIC), and would have found a solution whose objective value is 10% above the final optimum in only 9s (FCO) and 5s (VIC). Conversely, on infrastructure (i) (*i.e.*, using 64, *i.e.*, just 1), it would have required 2h2m36s (FCO) and 3h16m23s (VIC) to terminate, and 2m29s (FCO) and 22s (VIC) to get to 10% above the final optimum.

The time required by GD-Cover to estimate, via statistical model checking, the objective value yielded by each candidate deployment ranges within 1.8s–8.5s (FCO, 10–20 sensors) and 2s–32s (VIC 3–10 sensors), and grows with the number of pairs of sensors that might triangulate. The higher times measured in VIC (which has a smaller RoI than FCO, although with more obstacles) are indeed due to the fact that each sensor can in principle cooperate with most of the others. Hence, the number of pairs of sensors which might triangulate is much higher, and determining the coverage of each sampled point requires more effort. However, deploying GD-Cover using a high enough number of helper processes successfully mitigates this issue. Details are delayed to Appendix D-D. Finally, Appendix D-C evaluates the effectiveness of launching the parallel optimisers starting from the best initial deployments. This step greatly helps when in presence of tight time budgets and small computational infrastructures. Namely, for both case studies, the first deployment within 110% of the final optimum was achieved from a parallel NOMAD run starting from one of the 4 best random initial deployments, and the final optimum from one of the 20 best initial deployments (with one single exception).

*2) Computation of uncovered region in closed form:* This job is computationally way more intensive than estimating the objective value of a candidate deployment via statistical model checking. This is why such computation is performed only at the end of the optimisation process, or on user demand.

We ran GD-Cover to compute the uncovered region for the final (optimal) deployments. As explained in Section V-C, we enabled parallel computation also for this job, by splitting the two RoIs in different numbers of identical cells, which were processed in parallel by the available helper processes. A full scalability analysis for this job is delayed to Appendix D-B2. Here, we just mention that, differently from what happens during optimisation, the geometric reasoning required by the computation of the uncovered region is somewhat *hindered* when using large infrastructures. This is unsurprising, since processing too many small cells may yield duplication of efforts, because some of the computed polyhedra (those needed to represent the regions of Proposition V.1) would span a high number of cells. The optimal splitting of each RoI appears to be in a few thousands of cells. This yields the computation terminate in 31m33s (FCO, 6400 cells of size $100\times100\times10$ m$^3$) and 14m5s (VIC, 6498 cells of size $21\times21\times22$ m$^3$) on infrastructure (i) (64). By comparing these durations with those of the statistical model checking–based estimation of the objective value (which is also much more suitable to be massively parallelised), we see that driving the optimisation with such approximations is extremely beneficial.

Finally, Figure 6 shows, for each case study, the computed positions of the sensors and the uncovered region at three milestones of the optimisation process (when using the best number of sensors): the initial deployment (time zero), the first deployment whose objective value is 10% above the final optimum, and the final (optimal) deployment. The pictures are taken from our 3D visualiser.

## VII. Conclusions

In this article we proposed a novel approach to the computation of optimal (with respect to coverage, cost-effectiveness,

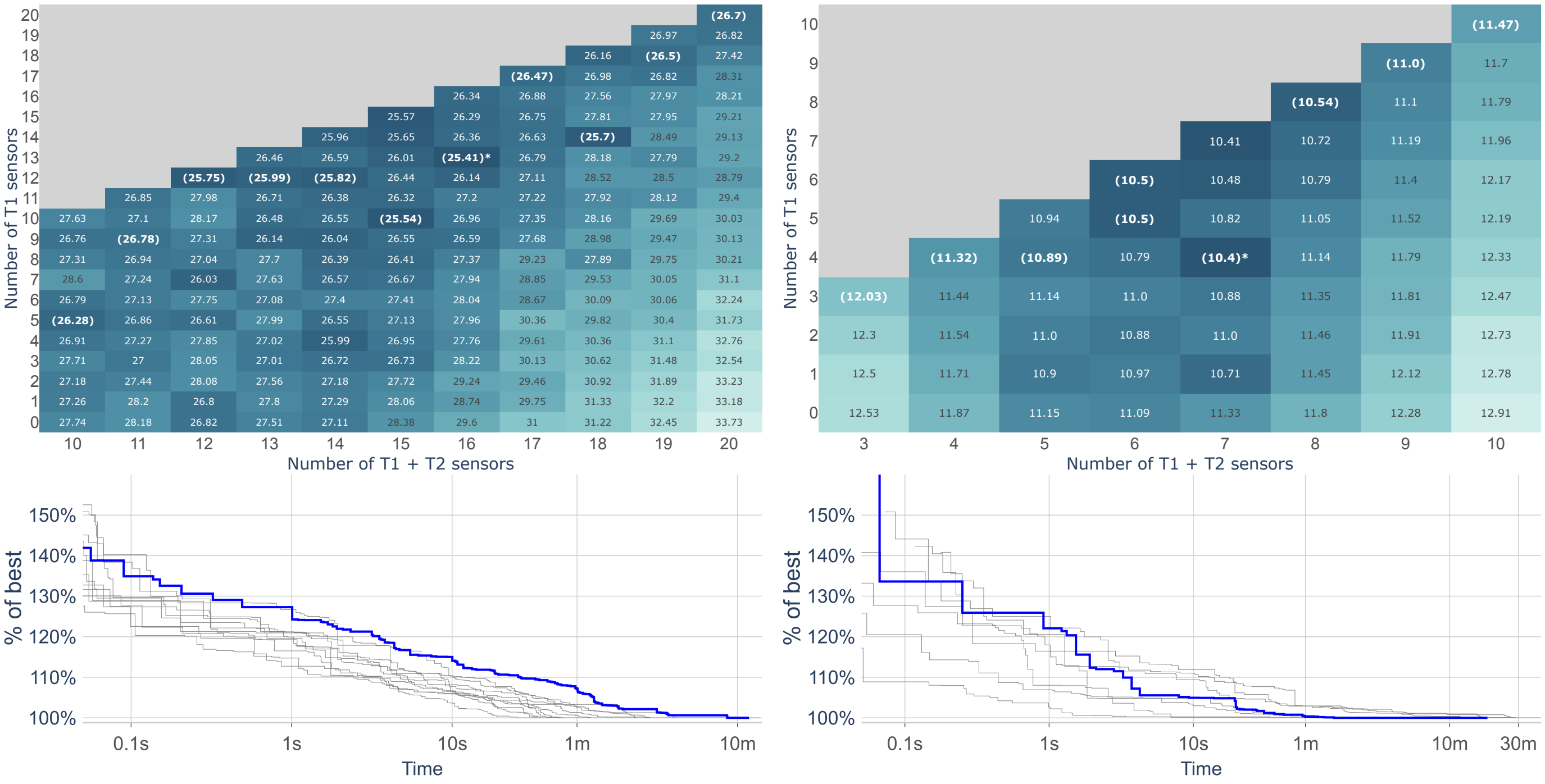

(a) Leonardo Da Vinci International Airport (FCO)

(b) Vienna International Center (VIC)

Figure 5: Up: best Overall Deployment Cost found for different numbers of T1 vs. T2 sensors (configurations). Down: Time course of the objective value during optimisation, when fully reserving 640 (*i.e.*, 10, infrastructure (ii) in Section VI-A5; one line per configuration; the blue lines refer to optimal configurations).

multiple sensing quality levels, and tolerance to sensor faults) deployments of triangulating sensors for unauthorised UAV localisation in large, complex critical 3D regions exhibiting obstacles (*e.g.*, buildings), varying terrain elevation, portions with different coverage priorities, and in presence of constraints on where sensors can actually be placed. Our approach relies on computational geometry and statistical model checking, effectively exploiting *off-the-shelf* AI-based BBO solvers, and enables the computation of a *closed-form, analytical* representation of the region uncovered by a sensor deployment, which provides the means for *rigorous, formal certification* of the quality of the latter. To our knowledge, no other method is available which addresses all such aspects (see also the recap in Appendix A).

We have demonstrated the practical feasibility of our approach by computing, in a *few minutes on a small parallel infrastructure* (or a few hours on a single workstation), optimal sensor deployments for UAV localisation in two large, complex regions, Leonardo Da Vinci International Airport (FCO) in Rome and Vienna International Center (VIC), using multiple instances of the NOMAD state-of-the-art AI-based BBO solver as the underlying optimisation engine.

Future work include the evaluation of other (*e.g.*, derivative-free) optimisation techniques and the improvement of GD-Cover to further reduce its computation time.

*Acknowledgments:* The authors are grateful to the anonymous reviewers as well as to G. Nazzaro and C. Giannetti for their support in implementing our 3D visualiser.

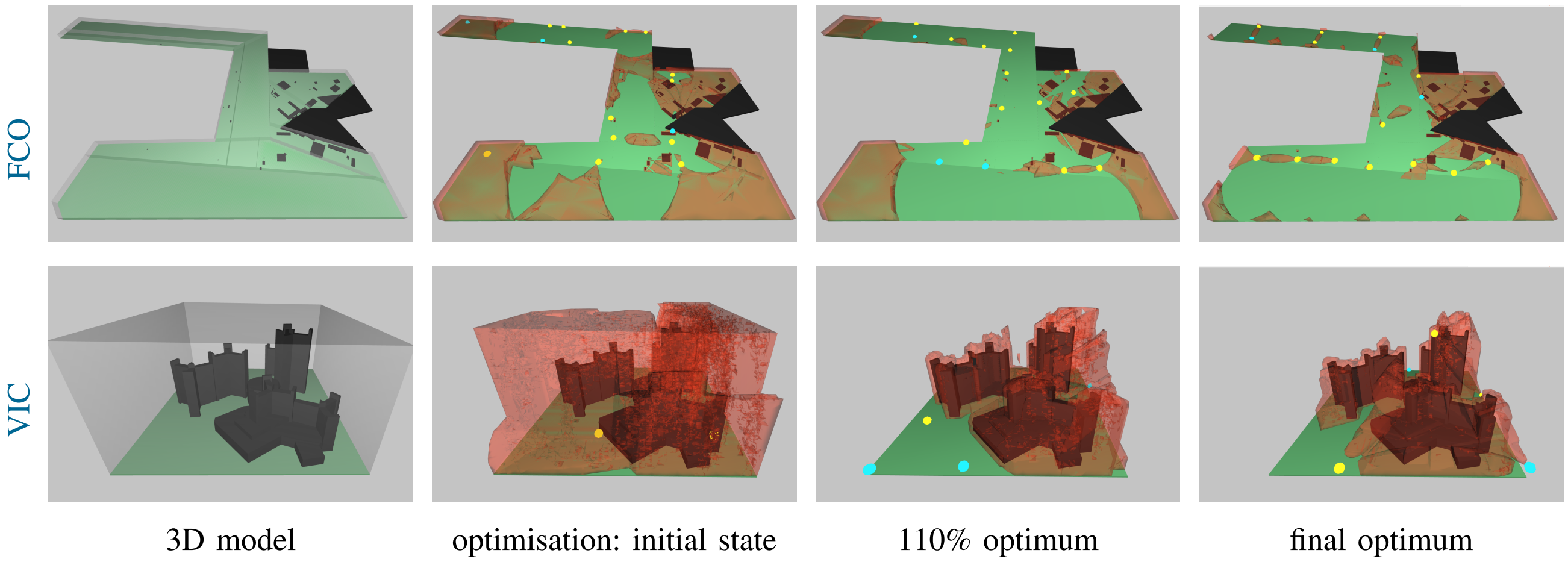

Figure 6: 3D models of our case studies and computation of optimal deployments at different states of search (green: RoI; red: uncovered region under no sensors faults).

# Appendix A
# Related work

In Table I we compare our approach against the related work discussed in Section II according to a number of benchmark criteria of interest, in order to better show the position of this article within the existing literature.

# Appendix B
# Proof of results

In this appendix we show the proofs of our results.

**Proposition V.1** (Geometric representation of the $(j,q)$-uncovered region)**.** The $(j,q)$-uncovered region of Definition III.2 can be equivalently defined as:

$$\mathcal{U}^{j,q} = \bigcup_{\substack{F \in 2^{\mathcal{S}} \\ |F| \leq j}} \bigcap_{\substack{(s_1,s_2) \in (\mathcal{S}-F)^2 \\ s_1 \neq s_2}} \mathcal{U}^q_{s_1,s_2} - \mathcal{O} \quad (7)$$

with

$$\mathcal{U}^q_{s_1,s_2} = \left[\mathcal{U}^{\text{range},q}_{s_1} \cup \mathcal{U}^{\text{range},q}_{s_2}\right] \cup \left[\mathcal{U}^{\mathcal{O},q}_{s_1} \cup \mathcal{U}^{\mathcal{O},q}_{s_2}\right] \cup \mathcal{U}^{\text{angle},q}_{s_1,s_2} \quad (8)$$

where, for $i \in [1,2]$:

- $\mathcal{U}^{\text{range},q}_{s_i}$ is the *region out of range* of $s_i$:

  $$\mathcal{U}^{\text{range},q}_{s_i} = \{X \in \mathcal{R} \mid \textit{dist}(X, \mathcal{D}(s_i)) > r_{s_i,q}\},$$

  *i.e.*, the complement of a sphere of radius $r_{s_i,q}$ centred in the position of sensor $s_i$, $\mathcal{D}(s_i)$.
- $\mathcal{U}^{\mathcal{O},q}_{s_i}$ is the region *not covered* by $s_i$ *because of obstacles*. It is the set of points $X \in \mathcal{R}$ such that the distance between an obstacle (a point in $\mathcal{O}$) and the straight-line segment connecting $X$ with $\mathcal{D}(s_i)$ is at most the First Fresnel Zone (FFZ) radius of $s_i$ for quality level $q$, $f_{s_i,q}$:

  $$\mathcal{U}^{\mathcal{O},q}_{s_i} = \textit{proj}(\mathcal{D}(s_i), \textit{bloat}(\mathcal{O}, f_{s_i,q})) \cap \mathcal{R}.$$

- $\mathcal{U}^{\text{angle},q}_{s_1,s_2}$ is the *region not covered* by $s_1$ and $s_2$ *because of excessive sensing error*. It is the set of points $X \in \mathcal{R}$ such that the angle $\angle\mathcal{D}(s_1)X\mathcal{D}(s_2)$ formed by $X$ with the positions of the two sensors is outside range $[\theta_{\min,q}, \theta_{\max,q}]$:

  $$\mathcal{U}^{\text{angle},q}_{s_1,s_2} = \{X \in \mathcal{R} \mid \angle\mathcal{D}(s_1)X\mathcal{D}(s_2) \notin [\theta_{\min,q}, \theta_{\max,q}]\}.$$

*Proof.* **($\mathbf{X} \in$ (1) $\implies \mathbf{X} \in$ (7))** By contraposition. Take any $X \notin$ (7). This means that either $X \notin \mathcal{R}$ or $X \in \mathcal{O}$ (in which cases, trivially $X \notin$ (1)), or, for all $F \in 2^{\mathcal{S}}$ such that $|F| \leq j$, there exist two distinct sensors $s_1$ and $s_2$ not in $F$ for which $X \notin \mathcal{U}^q_{s_1,s_2}$. The latter condition in turn means that all the conditions of Definition III.1 hold, namely:

- Since $X \notin \mathcal{U}^{\text{range},q}_{s_i}$, then $\textit{dist}(X, \mathcal{D}(s_i)) \leq r_{s_i,q}$, $i \in [1,2]$;
- Since $X \notin \mathcal{U}^{\mathcal{O},q}_{s_i}$ and $X \in \mathcal{R}$, then $X \notin \textit{proj}(\mathcal{D}(s_i), \textit{bloat}(\mathcal{O}, f_{s_i,q}))$, which means that $\textit{dist}(X\mathcal{D}(s_i), \mathcal{O}) > f_{s_i,q}$, $i \in [1,2]$;
- Since $X \notin \mathcal{U}^{\text{angle},q}_{s_1,s_2}$, then $\angle\mathcal{D}(s_1)X\mathcal{D}(s_2) \in [\theta_{\min,q}, \theta_{\max,q}]$.

Consequently, $\textit{cover}^q(X, s_1, s_2) = 1$ for all such pairs of sensors, hence $\textit{cover}^{j,q}(X) = 1$ (Definition III.2) and $X \notin$ (1).

**($\mathbf{X} \in$ (7) $\implies \mathbf{X} \in$ (1))** Again, by contraposition. Take any $X \notin$ (1). This means that either $X \notin \mathcal{R}$ or $X \in \mathcal{O}$ (in which cases, trivially $X \notin$ (7)), or $\textit{cover}^{j,q}(X) = 1$ (Definition III.2). The latter condition means that, for all $F \in 2^{\mathcal{S}}$ such that $|F| \leq j$, there exist two distinct sensors $s_1$ and $s_2$ not in $F$ for which $\textit{cover}^q(X, s_1, s_2) = 1$. This in turn means that $X$ satisfies all conditions of Definition III.1. Hence:

- Since $\textit{dist}(X, \mathcal{D}(s_i)) \leq r_{s_i,q}$, $i \in [1,2]$, then $X \notin \mathcal{U}^{\text{range},q}_{s_i}$;
- Since $\textit{dist}(X\mathcal{D}(s_i), \mathcal{O}) > f_{s_i,q}$, $i \in [1,2]$, then $X \notin \textit{proj}(\mathcal{D}(s_i), \textit{bloat}(\mathcal{O}, f_{s_i,q}))$, which, given that $X \in \mathcal{R}$, means that $X \notin \mathcal{U}^{\mathcal{O},q}_{s_i}$;
- Since $\angle\mathcal{D}(s_1)X\mathcal{D}(s_2) \in [\theta_{\min,q}, \theta_{\max,q}]$, then $X \notin \mathcal{U}^{\text{angle},q}_{s_1,s_2}$.

Consequently, $X \notin \mathcal{U}^q_{s_1,s_2}$. □

**Observation V.1.** For every $j \in [0,k], q \in \mathcal{Q}$, let $Z^{j,q}$ be a real-valued random variable defined on probability space $(\Omega^{j,q}, \mathcal{F}^{j,q}, \textit{Pr}^{j,q})$ (Definition V.1) as: $Z^{j,q}(\perp) = 0$; $Z^{j,q}(h) = V_{\mathcal{R}} \times w(j,q,h)$ (for $h \in \mathcal{H}$).

The value of the objective function (2) evaluated for deployment $\mathcal{D}$ is $\textit{Placem}(\mathcal{D}) + \sum_{j=0}^{k} \sum_{q \in \mathcal{Q}} \mathbb{E}(Z^{j,q})$, where $\mathbb{E}^{j,q}(Z)$ is the *expected value* of $Z^{j,q}$.

*Proof.* Let $V_{\mathcal{R}} = \textit{volume}(\mathcal{R})$ (see Definition V.1). For all $i \in [0,k], q \in \mathcal{Q}$, we have:

$$\mathbb{E}(Z^{j,q}) = \sum_{h \in \mathcal{H}} \left( \cancel{V_{\mathcal{R}}} \times w(j,q,h) \times \frac{\textit{volume}(\mathcal{U}^{j,q} \cap \mathcal{R}_h)}{\cancel{V_{\mathcal{R}}}} \right).$$

By expanding (4) in (2) we have:

$$\begin{aligned} &\textit{Placem}(\mathcal{D}) + \textit{Uncov}(\mathcal{D}) = \\ &\textit{Placem}(\mathcal{D}) + \textstyle\sum_{j=0}^{k} \sum_{q \in \mathcal{Q}} \sum_{h \in \mathcal{H}} w(j,q,h) \times \textit{volume}(\mathcal{U}^{j,q} \cap \mathcal{R}_h) = \\ &\textit{Placem}(\mathcal{D}) + \textstyle\sum_{j=0}^{k} \sum_{q \in \mathcal{Q}} \mathbb{E}(Z^{j,q}). \end{aligned}$$

□

# Appendix C
# Pseudo-code of GD-Cover

Here we give more details as well pseudo-code of our implementation of key components of Geometry-based Sensor Deployment Coverage Analyser (GD-Cover), our black box, namely the data structure implementing Axis-Aligned Bounding Box (AABB) tree–based indexed unions of polyhedra (Appendix C-A) and the function which computes polynomial approximations of the uncovered region (Section C-B).

## A. Indexed unions of polyhedra

Algorithm 1 defines our data structure implementing AABB tree–based Indexed Unions of Polyedra (iUoPs), using the algorithm from [24]. The data structure provides operations to create an iUoP, compute the intersection and the union of a collection of iUoPs, and the difference between two iUoPs. Function *simplify*() heuristically merges together two polyhedra in a iUoPs whose union is a polyhedron until a fix-point is reached.

| Approach | 3D RoI | Optimis. wrt. coverage | Optimis. wrt. cost | Sensors of different typologies | Sensors in 3D space | Sensor positioning constraints | 3D obstacles | Varying terrain elevation | Multiple coverage priorities | Multiple sensing quality levels | Fault tolerance | Uncovered region in closed form |
|---|---|---|---|---|---|---|---|---|---|---|---|---|
| This article | yes | yes | yes | yes | yes | yes | yes | yes | yes | yes | yes | yes |
| [1] | grid | yes | yes | yes | grid | yes | yes | yes | – | – | – | – |
| [2] | finite pts | yes | n. sensors | – | finite pts | yes | yes | yes | yes | – | – | – |
| [3], [4] | 3D surface | yes | yes | yes | 3D surface | yes | – | yes | yes | – | – | – |
| [5] | 3D surface | yes | – | – | 3D surface | – | – | yes | yes | – | – | – |
| [6] | – | yes | yes | – | – | 2D grid | – | – | – | – | – | – |
| [7] | – | yes | yes | – | – | finite pts | – | – | – | – | yes | – |
| [8], [9] | 3D surface | yes | yes | yes | 3D surface | yes | yes | yes | – | – | – | – |
| [10] | – | yes | n. sensors | yes | – | – | – | – | yes | – | – | – |
| [11] | – | yes | n. sensors | – | – | yes | – | – | – | – | – | – |
| [12] | – | yes | yes | yes | – | yes | – | – | – | – | – | – |
| [13] | – | yes | yes | yes | – | finite pts | – | – | – | – | – | – |
| [14] | – | yes | – | yes | – | yes | – | – | – | – | – | – |
| [15] | – | yes | n. sensors | – | yes | yes | – | – | – | – | – | – |
| [16] | 3D surface | yes | n. sensors | – | 3D surface | – | yes | yes | – | yes | – | – |
| [17] | – | yes | – | yes | yes | yes | – | – | yes | – | – | – |
| [18] | grid | yes | yes | yes | grid | yes | yes | yes | yes | – | – | – |
| [19] | grid | yes | n. sensors | yes | grid | yes | yes | yes | – | – | – | – |
| [20] | grid | yes | – | – | grid | yes | yes | yes | yes | – | – | – |
| [21] | finite pts | yes | – | – | yes | yes | – | – | – | – | – | – |
| [22] | grid | yes | – | – | – | – | yes | yes | – | – | – | – |
| [23] | grid | yes | yes | yes | – | yes | yes | yes | yes | – | – | – |

Table I: Comparison of our method against the other approaches discussed in Section II of the main article, over a set of benchmarking criteria.

Note: references in the table are numbered according to the References section of *this* (separate) supplementary material, and *do not* follow the numbering of references of the main article.

```
/* Instances of data structure iUoP define Indexed
   Unions of Polyedra, i.e., unions of polyhedra
   together with a spatial AABB tree index. iUoPs
   support efficient index-based intersection, union and
   difference operations.                                  */
data structure iUoP
    field polys, collection of closed convex polyhedra;
    field index, AABB tree; /* The leaves of the tree are
      the polyhedra in polys                                 */
function iUoP_create(polys)
    input polys, a collection of closed convex polyhedra;
    output a new iUoP;
    index ← build the AABB tree index using the
      minimum surface heuristics from [24];
    return iUoP(polys, index);
function iUoP_intersect(A, B)
    input A, B, two iUoPs;
    output a new iUoP representing A ∩ B;
    polys ← compute A.index ∩ B.index using the AABB
      tree intersection algorithm from [24]; returns a set
      of polyhedra;
    polys ← simplify(polys);
    return iUoP_create(polys);
function iUoP_union(L)
    input L, a collection of iUoPs;
    output a new iUoP representing ⋃_{u∈L} u;
    polys ← set of all polyhedra in all iUoPs of L;
    polys ← simplify(polys);
    return iUoP_create(polys);
function iUoP_diff(A, B)
    input A, B, two iUoPs;
    output a new iUoP representing A \ B;
    polys ← compute A \ B from A.index and B.index
      using the AABB tree intersection algorithm from
      [24] and complement; returns a set of polyhedra;
    polys ← simplify(polys);
    return iUoP_create(polys);
```

**Algorithm 1:** Data structure implementing AABB tree–based indexed unions of polyhedra.

### *B. Computing polyhedral approximations of the uncovered region*

Algorithm 2 shows pseudo-code of the GD-Cover function which computes a polyhedral over-approximation $\lceil \mathcal{U}^{j,q} \rceil$ of the region $(j,q)$-uncovered by the given sensor deployment $\mathcal{D}$ (the algorithm which computes a polyhedral under-approximation is similar).

Algorithm 2 is split into three phases. The first (sequential and quick) phase, just computes once and for all the unordered pairs of distinct sensors that are close enough to possibly interact with each other.

The second phase computes, in parallel exploiting the available helper processes, for all pairs $(s_1, s_2)$ of such sensors, the region $(j,q)$-uncovered by them ($\mathcal{U}_{s_1,s_2}$) and the region in the range of both of them ($\mathcal{R}_{s_1,s_2}$). This is done by running function *process_sensor_pair*$(s_1, s_2, q)$, described below, which computes (with the aid of the indexed data structure shown in Algorithm 1) the regions defined in Proposition V.1 of the main article. Such regions are broadcasted to all helper processes once and for all.

Finally, the third phase of the algorithm computes, again in parallel on the available helper processes, the portion $\lceil \mathcal{U}_c^{j,q} \rceil$ of each cell $c \in [1, m]$ $(j,q)$-uncovered by the input sensor deployment $\mathcal{D}$.

*Function process_sensor_pair():* For every unordered pair of sensors $(s_1, s_2)$ that are close enough to possibly interact with each other, function *process_sensor_pair*$(s_1, s_2, q)$ computes (with the aid of the indexed data structure shown in Algorithm 1) polyhedral under-/over-approximations of the regions defined in Proposition V.1 of the main article, *i.e.*, $\mathcal{U}_{s_i}^{\text{range},q}$, $\mathcal{U}_{s_i}^{\mathcal{O},q}$ ($i \in \{1,2\}$), and $\mathcal{U}_{s_1,s_2}^{\text{angle},q}$. Every region is under-/over-approximated with a maximum tolerance of $\rho \in \mathbb{R}_+$, a user parameter.

Here we briefly outline how this function behaves.

A core sub-problem when computing polyhedral under-/over-approximations of each of the required regions is computing a polyhedral under-/over-approximation of a sphere. Given a sphere $S$ having radius $r$, a polyhedral under- (respectively over-) approximation $\lfloor S \rfloor$ (respectively $\lceil S \rceil$) of $S$ with tolerance $\rho$ can be computed as a *geodesic polyhedron* inscribed in $S$ (respectively circumscribing $S$). The number of vertices of such (regular) polyhedra can be derived from the radius of $S$ and the required tolerance $\rho$ using standard geometry arguments. The reader is referred to, *e.g.*, [25], [26] for details.

Now, as for the regions of Proposition V.1:

*a) Region $\mathcal{U}_{s_i}^{\text{range},q}$:* This is the complement of a sphere $S$ of radius $r_{s_i,q}$ centred in $\mathcal{D}(s_i)$.

Thus, $\lfloor \mathcal{U}_{s_1}^{\text{range},q} \rfloor$ and $\lceil \mathcal{U}_{s_1}^{\text{range},q} \rceil$ are respectively computed as $\mathcal{R} - \lceil S \rceil$ and $\mathcal{R} - \lfloor S \rfloor$. As explained in the main paper, although the difference between two unions of polyhedra is a union of non-closed polyhedra, we can safely convert it into a union of closed polyhedra, given that tolerance $\rho$ is strictly positive.

*b) Region $\mathcal{U}_{s_i}^{\mathcal{O},q}$:* This is the intersection of $\mathcal{R}$ with the projection of point $\mathcal{D}(s_i)$ onto *bloat*$(\mathcal{O}, f_{s_i,q})$.

A $f_{s_i,q}$-bloating of a union of polyhedra $\mathcal{O}$ can be obtained as the union of $f_{s_i,q}$-bloatings of each polyhedron $O \in \mathcal{O}$. Also, the projection of point $X$ onto a union of regions is the union of projections of $X$ onto each such region. Thus, the core problems to be solved here are how to compute polyhedral under- and over-approximation of a $f_{s_i,q}$-bloating of a single polyhedron. How to compute projections is described elsewhere [27].

It is immediate to see that a possible $f_{s_i,q}$-bloating of a polyhedron $O$ can be simply defined as:

$$\textit{bloat}(O, f_{s_i,q}) = \textit{conv}(O \cup \{S_v \mid v \text{ is a vertex of } O\}) \quad (9)$$

where $S_v$ is the sphere centred in $v$ having radius $f_{s_i,q}$ and *conv*$(R)$ is the convex-hull of region $R$ (see Figure 7 for an example in 2D). Note that the computed convex-hull is not **function** *poly_over*($\mathcal{D}$, $j$, $q$)
  **param** $\mathcal{R}$, *Region of Interest*;
  **param** *cells*($\mathcal{R}$), *a iUoP defining a cell-wise partitioning of RoI (for parallel computation)*;
  **param** $\mathcal{H}$, *priorities*;
  **param** $\mathcal{O}$, *obstacles*;
  **param** $\mathcal{S}$, *sensors and their properties*;
  **input** $\mathcal{D}$, *a deployment of sensors in* $\mathcal{S}$;
  **input** $j$, *max n. of sensor faults*;
  **input** $q$, *sensing quality level*;
  **output** $\lceil \mathcal{U}^{j,q} \rceil$, *polyhedral over-approximation of the region* $(j, q)$*-uncovered by* $\mathcal{D}$;
  /* Phase 1, initialise maps:
  • sensor_pairs, pairs of sensors close enough to possibly interact with each other at quality level q
  • sensor_pairs_U[sensor pair], region (as iUoP) uncovered by 'sensor pair' for quality level q
  • sensor_pairs_R[sensor pair], region (as iUoP) within the reach of both sensors or 'sensor pair' for q */
  *sensor_pairs* ← unord'd pairs of distinct sensors $\{s_1, s_2\}$ s.t. $dist(\mathcal{D}(s_1), \mathcal{D}(s_2)) < r_{s_1,q} + r_{s_2,q}$;
  *sensor_pairs_U* ← empty map;
  *sensor_pairs_R* ← empty map;
  /* Phase 2 */
  **parallel foreach** $\{s_1, s_2\} \in$ *sensor_pairs* **do**
    /* parallel computation on helper processes via the centralised task dispatcher */
    $(\mathcal{U}_{s_1,s_2}, \mathcal{R}_{s_1,s_2}) \leftarrow$ *process_sensor_pair*$(s_1, s_2, q)$;
    **if** $\mathcal{R} \subseteq \mathcal{U}_{s_1,s_2}$ **then**
      remove $\{s_1, s_2\}$ from *sensor_pairs*;
    **else**
      *sensor_pairs_U*$[\{s_1, s_2\}] \leftarrow \mathcal{U}_{s_1,s_2}$;
      *sensor_pairs_R*$[\{s_1, s_2\}] \leftarrow \mathcal{R}_{s_1,s_2}$;
  **send** *sensor_pairs*, *sensor_pairs_U*, *sensor_pairs_R* to helper processes;
  /* Phase 3 */
  $\lceil \mathcal{U}^{j,q} \rceil \leftarrow$ empty set of polyhedra;
  **parallel foreach** $c \in$ *cells*($\mathcal{R}$) **do**
    /* parallel computation on helper processes via the centralised task dispatcher */
    $\lceil \mathcal{U}_c^{j,q} \rceil \leftarrow$ *poly_over_in_cell*($c$, $j$, $q$);
    $\lceil \mathcal{U}^{j,q} \rceil \leftarrow$ iUoP *_union*$(\lceil \mathcal{U}^{j,q} \rceil, \lceil \mathcal{U}_c^{j,q} \rceil)$;
  **return** $\lceil \mathcal{U}^{j,q} \rceil$;

**Algorithm 2:** Computation of polyhedral over-approximation $\lceil \mathcal{U}^{j,q} \rceil$ of the $(j, q)$-uncovered region.

**function** *poly_over_in_cell*($\mathcal{D}$, $c$, $j$, $q$)
  **global** *sensor_pairs*;
  **global** *sensor_pairs_U*;
  **global** *sensor_pairs_R*;
  **param** $\mathcal{S}$, *sensors and their properties*;
  **param** $\mathcal{R}$, *Region of Interest*;
  **input** $\mathcal{D}$, *a deployment of sensors in* $\mathcal{S}$;
  **input** $c$, *a cell of* $\mathcal{R}$;
  **input** $j$, *max n. of sensor faults*;
  **input** $q$, *sensing quality level*;
  **foreach** $F \in 2^{\mathcal{S}}$ *s.t.* $|F| = j$ **do**
    *sensors_pairs_c* $\leftarrow \{\{s_1, s_2\} \in$ *sensor_pairs* $\mid s_1 \notin F,\ s_2 \notin F,$ iUoP *_intersect*$(c, sensor_pairs_R[\{s_1, s_2\}]) \neq \emptyset\}$;
    /* Compute $\lceil \mathcal{U}_{c,F}^{j,q} \rceil$, the region (iUoP) uncovered by all pairs of sensors not in $F$ */
    $\lceil \mathcal{U}_{c,F}^{j,q} \rceil \leftarrow c$;
    **foreach** $\{s_1, s_2\} \in$ *sensors_pairs_c* **do**
      $\lceil \mathcal{U}_{c,F}^{j,q} \rceil \leftarrow$ iUoP *_intersect*$(\lceil \mathcal{U}_{c,F}^{j,q} \rceil,$ *sensor_pairs_U*$[\{s_1, s_2\}])$;
      **if** $\lceil \mathcal{U}_{c,F}^{j,q} \rceil = \emptyset$ **then break**;
    $\lceil \mathcal{U}_c^{j,q} \rceil \leftarrow$ iUoP *_union*$(\lceil \mathcal{U}_c^{j,q} \rceil, \lceil \mathcal{U}_{c,F}^{j,q} \rceil)$;
    **if** $\lceil \mathcal{U}_c^{j,q} \rceil = c$ **then**
      **break** ; /* no need to consider other $F$s */
  **return** $\lceil \mathcal{U}_c^{j,q} \rceil$;

**Algorithm 3:** Computation of polyhedral over-approximation of the $(j, q)$-uncovered region in cell $c$ (helper process).

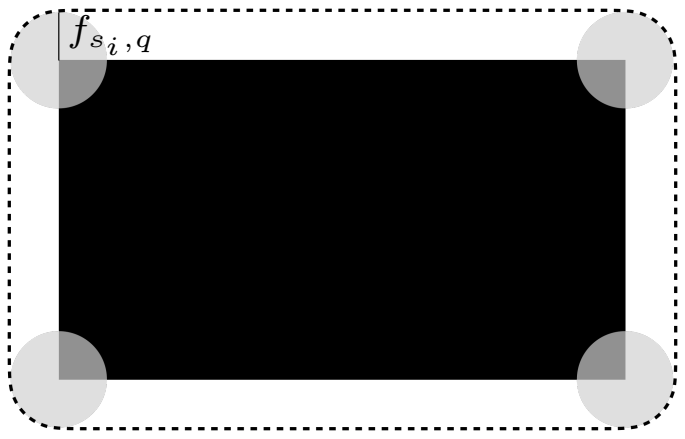

Figure 7: $f_{s_i,q}$-bloating of a polyhedron $O$ (2D, in black) as defined in (9).

a polyhedron, because of the presence of residual portions of the spheres $S_v$ outside $O$.

A polyhedral under- (respectively over-) approximation of *bloat*$(O, f_{s_i,q})$ (with tolerance $\rho$) can thus be computed as the convex-hull of the union of $O$ and $\lfloor S_v \rfloor$ (respectively $\lceil S_v \rceil$) for every vertex $v$ of $O$ (where the polyhedral under- and over-approximations of the spheres are computed with tolerance $\rho$).

*c) Region* $\mathcal{U}_{s_1,s_2}^{\text{angle},q}$*:* This is the set of points $X \in \mathcal{R}$ such that the angle $\angle \mathcal{D}(s_1) X \mathcal{D}(s_2)$ formed by $X$ with the positions of the two sensors is outside range $[\theta_{\text{min},q}, \theta_{\text{max},q}]$ and can be defined as the union of two regions:

a) $\{X \in \mathcal{R} \mid \angle \mathcal{D}(s_1) X \mathcal{D}(s_2) < \theta_{\text{min}}\}$ and
b) $\{X \in \mathcal{R} \mid \angle \mathcal{D}(s_1) X \mathcal{D}(s_2) > \theta_{\text{max}}\}$.

Region a) can be computed as follows. Consider the segment $\mathcal{D}(s_1)\mathcal{D}(s_1)$ connecting $s_1$ and $s_2$, any plane $\pi$ containing $\mathcal{D}(s_1)\mathcal{D}(s_1)$, and the locus of points $X \in \pi$ such that $\theta_{\min} \leq \angle\mathcal{D}(s_1)X\mathcal{D}(s_2)$. From classical results in geometry, the latter is a segment $\sigma$ of a circle having $\mathcal{D}(s_1)\mathcal{D}(s_1)$ as a chord (the white area in Figure 8a). Region a) is the complement of the solid of revolution obtained by rotating $\sigma$ around $\mathcal{D}(s_1)\mathcal{D}(s_2)$, as shown in Figure 8a. The region can be both under- and over-approximated (with tolerance $\rho$) as a union of polyhedra with similar methods as above.

Conversely, region b) can be computed by considering the circular segment defined by the locus of points $X \in \pi$ (with $\pi$ being any plane containing $\mathcal{D}(s_1)\mathcal{D}(s_1)$) such that $\theta_{\max} \leq \angle\mathcal{D}(s_1)X\mathcal{D}(s_2)$ (the grey area in Figure 8b). Region b) is the solid of revolution obtained by rotating such a circular segment around $\mathcal{D}(s_1)\mathcal{D}(s_2)$. Also this region can be both under- and over-approximated (with tolerance $\rho$) as a union of polyhedra with similar methods.

# Appendix D
# Experiments

Here we give more details on our experimental setting and on our experimental results.

## A. Experimental setting

*1) Running experiments on parallel infrastructures of various sizes:* A direct way of performing a throughout analysis of the performance of our system, also considering parallel infrastructures of various sizes, would have required us to run all our experiments (*i.e.*, for each of our case studies, for all combinations of T1/T2 sensors and all combinations for $n$ and $n_1$) multiple times, once for each candidate infrastructure. Each time, we would have needed to *fully reserve* the requested number of nodes *exclusively* to our job.

Such a direct approach would have been *unviable*, given that we were using a computational cluster *shared* among a high number of competing processes, because of the implemented jobs scheduling policies.

Hence, we proceeded in an indirect way. Namely, we ran our experiments *asynchronously*, using the meta-scheduler provided by the cluster in use (DAGMAN, https://htcondor.org/dagman/dagman.html). This means that each iteration of each process of Figure 3 of the main article (the Orchestrator, each of the $n_1$ Black-Box Optimisation, BBO, solvers, the GD-Cover tasks dispatcher, and each of the $n_2$ GD-Cover helpers) were run independently. Here, the term *iteration* refers to the time-span of a process starting from the reception of a message requesting a computation, and terminating with the sending back of the result of the requested computation.

Exchange of network messages were replaced by the sending process writing files in shared storage and the receiving process reading them back, and running times of all process iterations were precisely measured (excluding the I/O time due to file operations) and stored in log files. The time that would have been required for the exchange of real network messages for each needed type and payload size was measured through an independent experiment, by running a mock-up software implemented on top of OpenMPI (https://www.open-mpi.org), which exchanges a high number of messages of the required types and payload sizes (100000 messages for each type and each payload size), and computes the average *half turnaround time* for each of them.

Building on the generated log files containing the completion time and maximum memory occupation of each process iteration and the average network time for each envisioned message, we reconstructed what would have been the completion time of our system under the assumption that it were when launched on a *fully reserved* infrastructure of any given number of nodes (with nodes being identical to those of our cluster). This effectively emulates what would have happened in a computational infrastructure fully devoted to run our system.

Despite the fact that the completion times estimated in this way might be subject to (slight) approximations, results below on the scalability of our software under different business scenarios show speed-ups so substantial (from hours to a few minutes) that can easily incorporate any reasonable estimation error.

*2) Objective function weights:* Table II shows the values we used in our experiments for the weights $w(j, q, h)$ of the objective function in our two case studies: Leonardo Da Vinci International Airport (FCO) and Vienna International Center (VIC). They were estimated by preliminary experiments with the goal to achieve the best overall deployment using sensors of both types.

Note that, as for FCO, we set to 0 the weights representing the cost of not covering a unit of volume of the low priority region under $j = 1$ sensor fault. This is reasonable, given that the low priority region for that case study is extremely large (as it includes the airport runways).

## B. Scalability analysis of the parallel system

Here we perform a deeper analysis of the scalability of our parallel system to search for an optimal sensor deployment and to compute a closed-form polyhedral representation of the region uncovered by the latter.

*1) Optimisation:* Figures 9 and 10 show, for each case study and for different values for $n$ (total computational nodes) and $n_1$ (parallel optimisers), a figure similar to Figure 5 (down), that is, the time course of the objective value (normalised with respect to the value of the final optimum) of the best deployment found at time $t$. For each value of $n$, the boxed cell denotes the number of parallel optimisers ($n_1$) which yield maximum efficiency with respect to the sequential algorithm (conventionally defined as $n = n_1 = 1$, with neither orchestrator nor a dispatcher process).

It can be seen that, for both case studies, running our system on a single workstation ($n = 64$) allows it to terminate within a few hours. Completion time can be reduced to just a few minutes when deploying our system to a larger infrastructure ($n = 3200$, *i.e.*, 50 64-core machines), with efficiency of 72% (FCO) and 91% (VIC), noticeably higher in the latter more complex scenario.

Clearly, the completion times estimated above, computed as described in Appendix D-A1, might be subject to approximations. However, beyond being the only technically viable

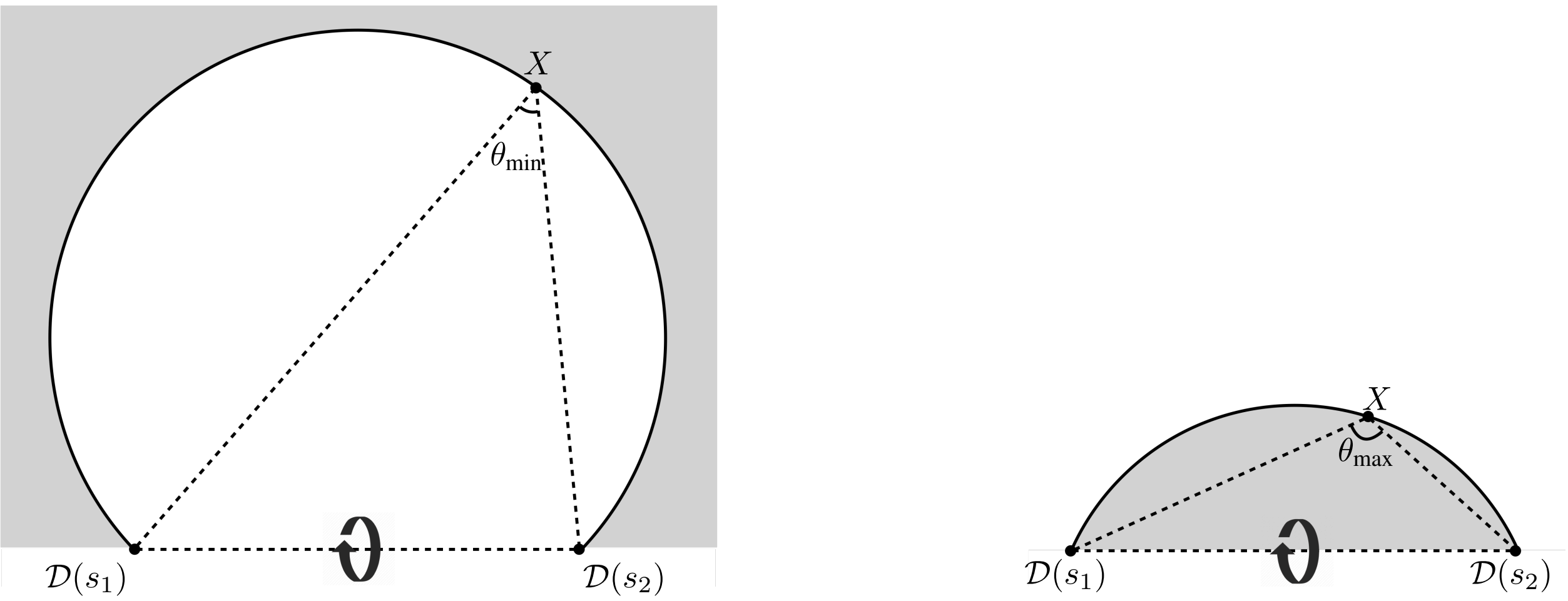

(a) $\{X \in \mathcal{R} \mid \angle\mathcal{D}(s_1)X\mathcal{D}(s_2) < \theta_{\text{min}}\}$ (grey part, complement of solid of revolution)

(b) $\{X \in \mathcal{R} \mid \angle\mathcal{D}(s_1)X\mathcal{D}(s_2) > \theta_{\text{max}}\}$ (grey part, solid of revolution)

Figure 8: Computation of region $\mathcal{U}^{\text{angle},q}_{s_1,s_2}$.

| faults ($j$) | quality ($q$) | priority ($h$) | $w(j,q,h)$ |
|---|---|---|---|
| 0 | $q_0$ | low | 10 |
| 0 | $q_0$ | high | 15 |
| 0 | $q_1$ | low | 15 |
| 0 | $q_1$ | high | 20 |
| 1 | $q_0$ | low | 0 |
| 1 | $q_0$ | high | 1 |
| 1 | $q_1$ | low | 0 |
| 1 | $q_1$ | high | 1 |

FCO

| faults ($j$) | quality ($q$) | priority ($h$) | $w(j,q,h)$ |
|---|---|---|---|
| 0 | $q_0$ | low | 5 |
| 0 | $q_0$ | high | 7 |
| 1 | $q_0$ | low | 5 |
| 1 | $q_0$ | high | 7 |
| 0 | $q_1$ | low | 0.5 |
| 0 | $q_1$ | high | 1 |
| 1 | $q_1$ | low | 1 |
| 1 | $q_1$ | high | 2 |

VIC

Table II: Values for weights $w(j,q,h)$ of the objective function used in our experiments.

solution applicable to the cluster at our disposal, they are accurate enough for our goal, *i.e.*, to give full evidence of the scalability of our software under different business scenarios, and in particular, that the computation can be easily sped-up from hours (on a single workstation) to just a few minutes using a small parallel infrastructure.

*2) Computation of the uncovered region:* Figure 11 shows the time required by GD-Cover to compute the closed-form polyhedral representation of the region uncovered by the best deployment found (under no sensor faults), for varying numbers of identical cells in which the RoI is split and for varying numbers of computational nodes.

It can be observed that, differently from what happens during optimisation, the geometric (symbolic) reasoning required by the computation of the uncovered region is somewhat *hindered* when using large infrastructures. This is unsurprising, since two different phenomena occur when parallelising this computation on a large number of nodes, and both of them obstruct such a large parallelisation:

(i) The processing time of each cell is largely variable. Namely, for those cells completely covered or completely uncovered by the deployment, the computation is very fast, while for those cells only partially covered, the computation is way more complex. This implies that, in order to achieve a good load balancing, the RoI must be split in a number of cells much higher than the number of available nodes.

(ii) With a higher number of (smaller) cells, greater duplication of efforts arises. This is because the probability that polyhedra to be computed (those needed to represent the regions of Proposition V.1) span a higher number of cells increases. Such polyhedra need to be computed for each cell they intersect.

### C. Effectiveness of sorting random initial deployments

Here we show that sorting the $N = 100$ initial random sensor deployments in descending order of their quality is a very effective heuristics for this problem, and greatly helps when in presence of tight time budgets and small computational infrastructures which would force us to use a number of parallel optimisers ($n_1$) much smaller than $N$. In particular, Figure 12 shows, for each case study, the distribution, among all configurations for that case study (*i.e.*, for all numbers of T1 and T2 sensors) of the number $i \in [1, N]$ of the initial random deployment (after sorting) which yielded, in the shortest time, a deployment within any given percentage of the final optimum. Interestingly, for both case studies, the first deployment within 120% of the final optimum was achieved from the parallel NOMAD run starting from the best random initial deployment; the first deployment within 110% of the final optimum was achieved from a parallel NOMAD run starting from one of the 4 best random initial deployments, and the final optimum was first computed from a parallel NOMAD run starting from one of the 20 best random initial deployments (with one single exception). This means that we could have safely removed, respectively, 99, 96 and 80 out of the $N = 100$ initial deployments from the sorted list and still get the same deployments within 120%, 110%, 100% of the final optimum (with one exception). Such computations would have been made much faster on the same computational infrastructure or, alternatively, would have been feasible in comparable time on a much smaller (hence, much cheaper) infrastructure.

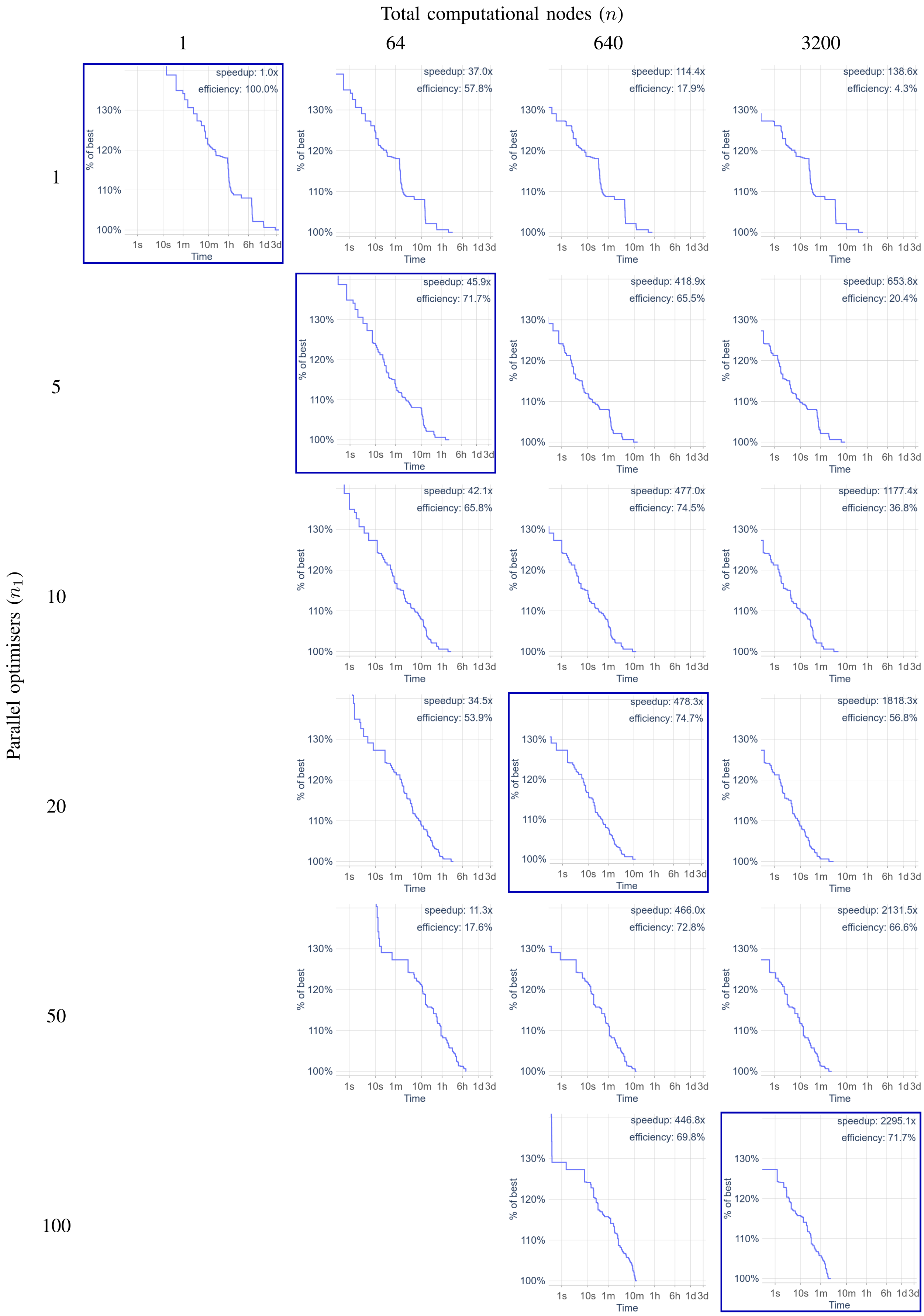

Figure 9: Leonardo Da Vinci International Airport (FCO): Time course of the (normalised) objective value of the best deployment found during optimisation, for varying numbers of parallel optimisers (rows) and overall computational nodes (columns).

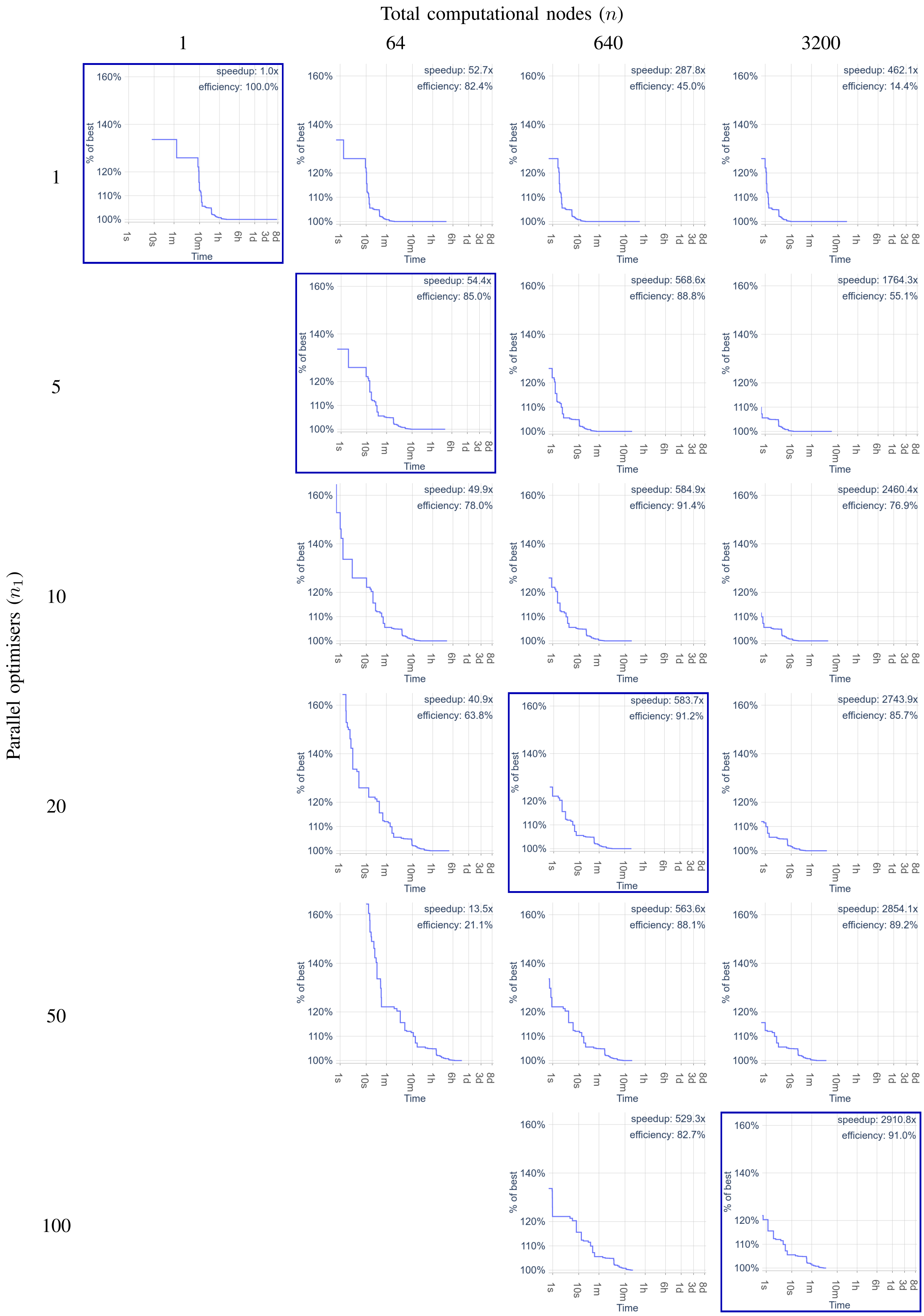

Figure 10: Vienna International Center (VIC): Time course of the (normalised) objective value of the best deployment found during optimisation, for varying numbers of parallel optimisers (rows) and overall computational nodes (columns).

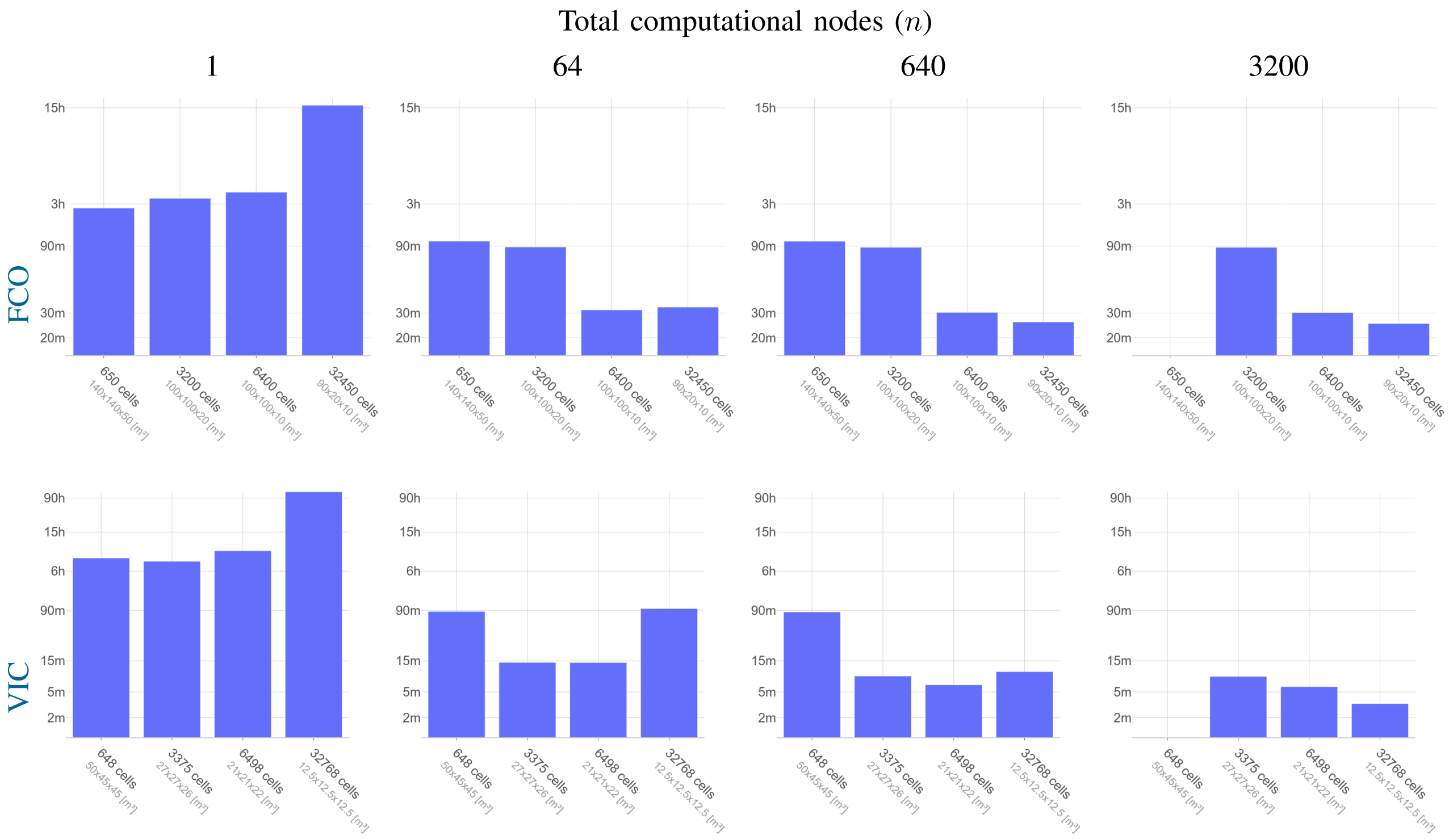

Figure 11: Time (log-scale) to compute the uncovered region in closed form for varying numbers of identical cells in which the RoI is split and for varying numbers of computational nodes.

### *D. Performance of the statistical model checking–based evaluation of a candidate deployment*

Figure 13 shows, for each case study, the average time required by GD-Cover to estimate, via statistical model checking, the objective value yielded by a candidate deployment, for each number of sensors.

The difference between the two case studies is due to intrinsic characteristics of the RoIs themselves. In fact, in FCO, GD-Cover running times are positively influenced by the fact that sensors are spread across a large space and each sensor has hope to cooperate only with few other sensors. This benefits GD-Cover when determining whether a point is covered by at least a pair of sensors. Conversely, in VIC (which has a smaller RoI, although with more obstacles), each sensor can in principle cooperate with most of the others. Hence, the number of pairs of sensors which might triangulate is higher, and determining the coverage of each sampled point requires more effort. Deploying GD-Cover using a high enough number of helpers successfully mitigates this issue.

## Appendix E
## Uncovered region 3D visualiser

Here we briefly showcase the proof-of-concept web application we developed for the 3D visualisation and interactive exploration of the regions uncovered by a sensor deployment.

The visualiser runs in any browser and has been realised with the *Three.js* Javascript framework.

Users of the application are initially prompted to upload one or more of the following inputs (in JSON format):

1) definition of the RoI, including a partition thereof in priority regions, obstacles, and the region where sensors can be deployed
2) a sensor deployment
3) the region not covered by the sensor deployment
4) the region not covered by each pair of cooperating deployed sensors.

All regions are given as unions of convex polyhedra. Since each input to the visualiser is either an input or an output of GD-Cover, all such data can be immediately obtained from GD-Cover itself, by instructing it to dump the requested data into one or more JSON files.

For reviewing purposes, a demo version of the application (already fed with the optimal deployments for our two case studies shown in Section VI-C) has been deployed at:

https://mclabservices.di.uniroma1.it/antidrone-visualizer/

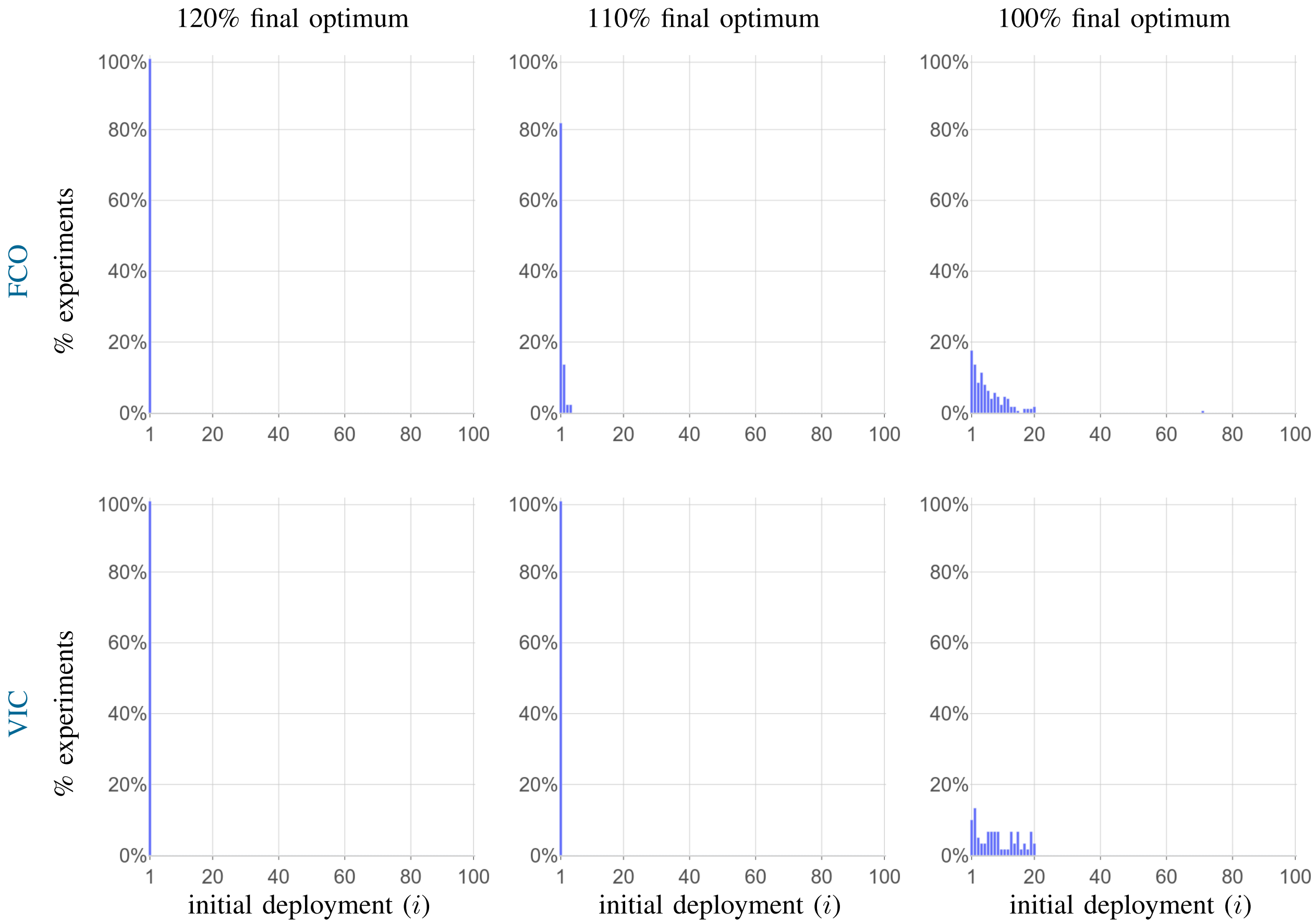

Figure 12: Distribution, among all experiments of each case study, of the number $i \in [1, N]$ of the initial random deployment (after sorting from the best to the worst) which yielded, in the shortest time, a deployment within any given percentage of the final optimum (minimum ODC).

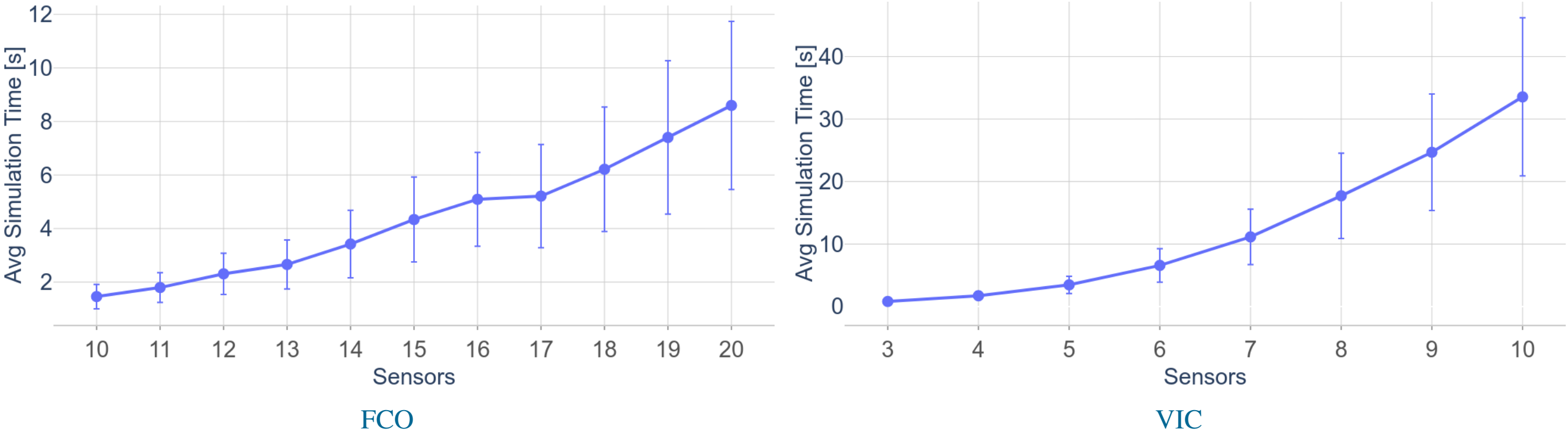

Figure 13: Average (plus min, max) time required by GD-Cover to estimate, via statistical model checking, the objective value yielded by a candidate deployment, for each number of sensors.

Figure 6 in the main article shows our two case studies, FCO and VIC, as visualised in the web application.

The user can see the region where sensors can be placed by pressing the "Admissible Placement Region" button in the side menu. Figure 14 shows the admissible placement region of T2 sensors in VIC. It can be seen that sensors can be deployed on poles on the ground (at a height of up to 15 m), over some of the roofs and on the concrete walls of the towers (on poles of length 5–10 m), but not on the glass façades.

Figure 15 shows the region with high coverage priority in FCO, which includes the runways and the areas around them. Users can enable and disable the visualisation of the region at each level $X$ of priority by pressing the "Region with Priority $X$" buttons in the side menu.

By selecting a pair of sensors via the "Sensors" drop-down menu, the user can see the region not covered by a specific pair of sensors. As an example, Figure 16 shows a detail of the region not covered by a specific pair of sensors (number 13 and 12) in FCO. The figure clearly highlights the portions of the RoI not covered by that pair of sensors due to each condition of Definition III.1 and Proposition V.1 in the main article (graphically shown in Figure 4 of the main article), namely: range of sensors, presence of obstacles, unsatisfactory sensing angle. Note that, being the left sensor placed at 14 m of height and the building closest to it is only 10 m high, the whole resulting projection intercepts the ground within a finite distance. By navigating the uncovered region in 3D, the user can easily assess the contribution of each pair of sensors to the coverage of the RoI.

Figure 17 displays, in its entirety, the region of VIC uncovered by the 4 T1 (cyan) and 3 T2 sensors (yellow) deployed as shown.

Finally, Figure 18 shows how our visualiser may be also employed to perform other kinds of what-if analyses on a computed deployment of sensors. As an example, we could be interested to see what would be the uncovered region if certain, specific sensors actually experience a fault. In particular, the figure shows the FCO and VIC regions that *would* remain uncovered by the respective optimal deployments of sensors (those shown in Figure 6 of the main article), in case the occurring sensor fault is such to maximise the ODC (*worst case*).

$LN@col2LN@col1$

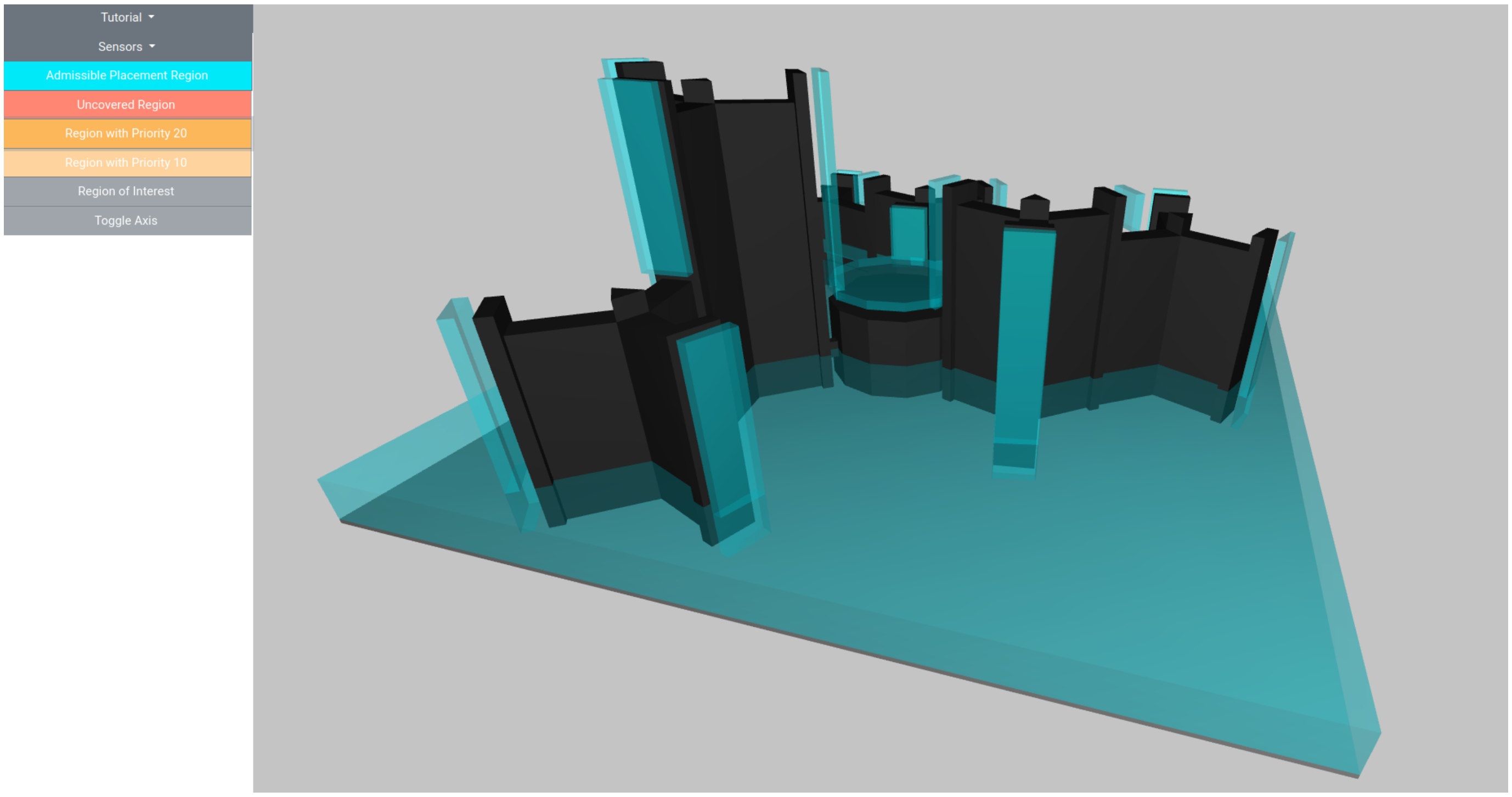

Figure 14: Admissible region of T2 sensors in VIC.

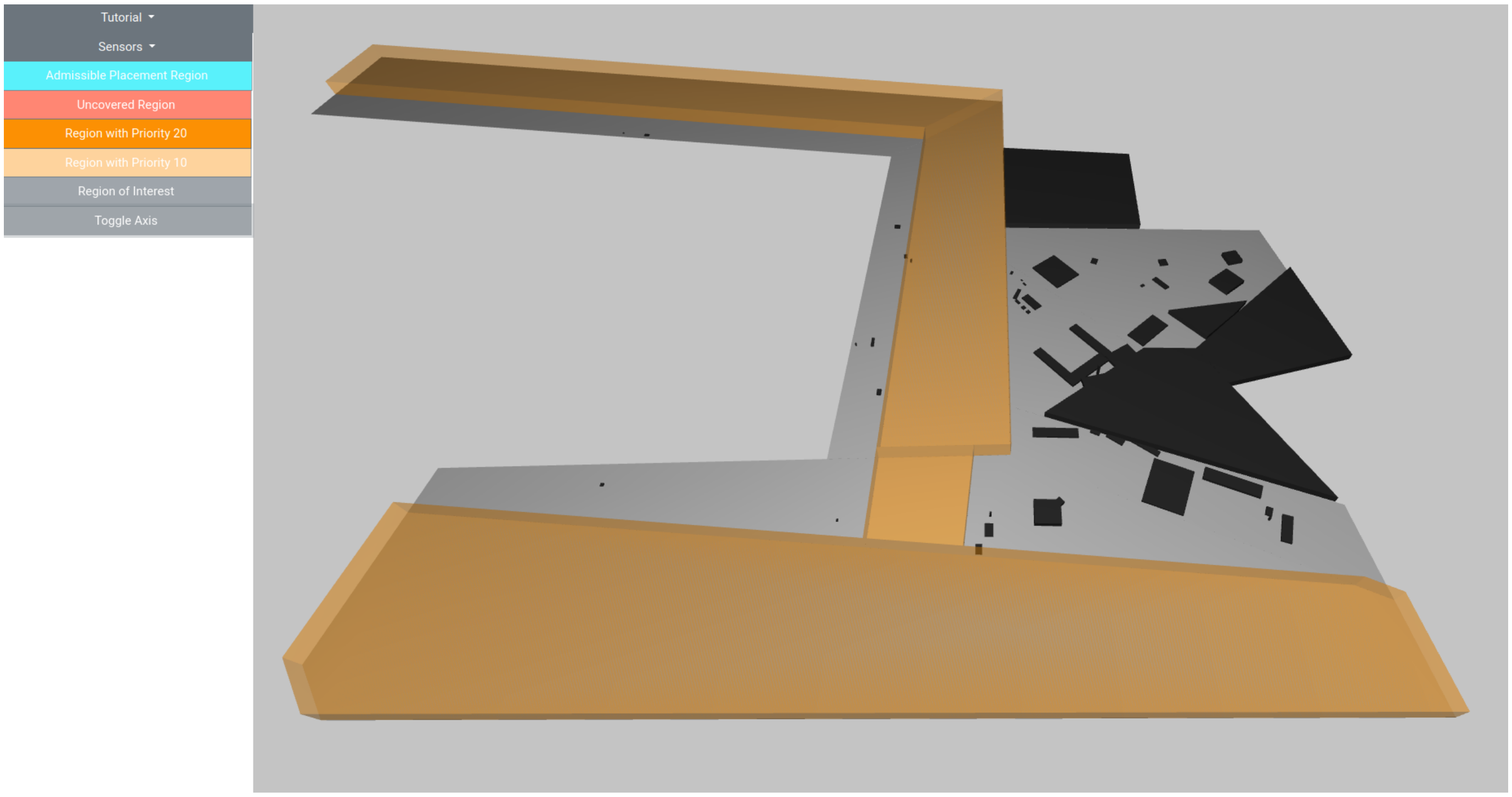

Figure 15: High-priority region in FCO.

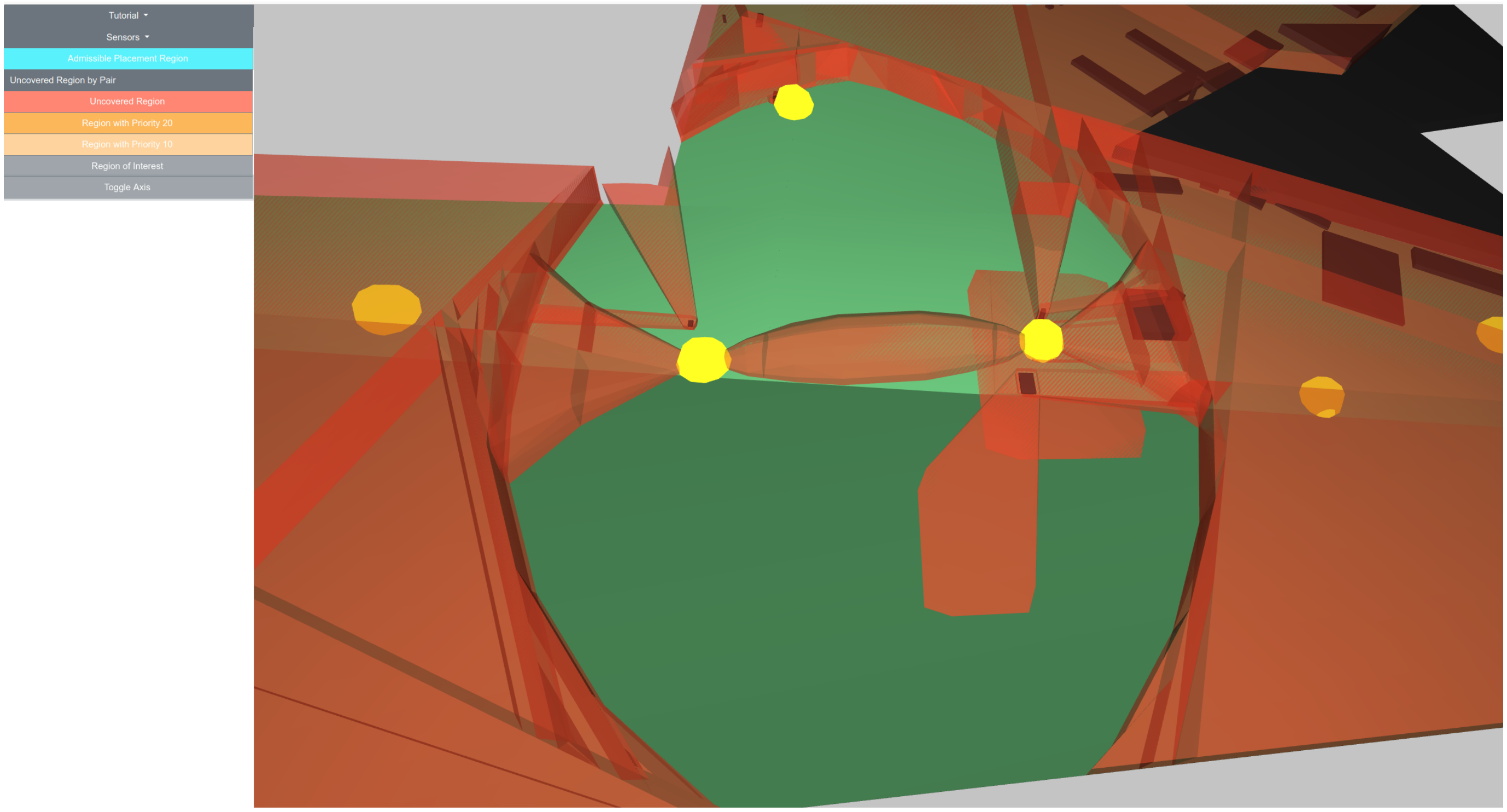

Figure 16: Details of the region not covered by sensors 13 and 12 in FCO.

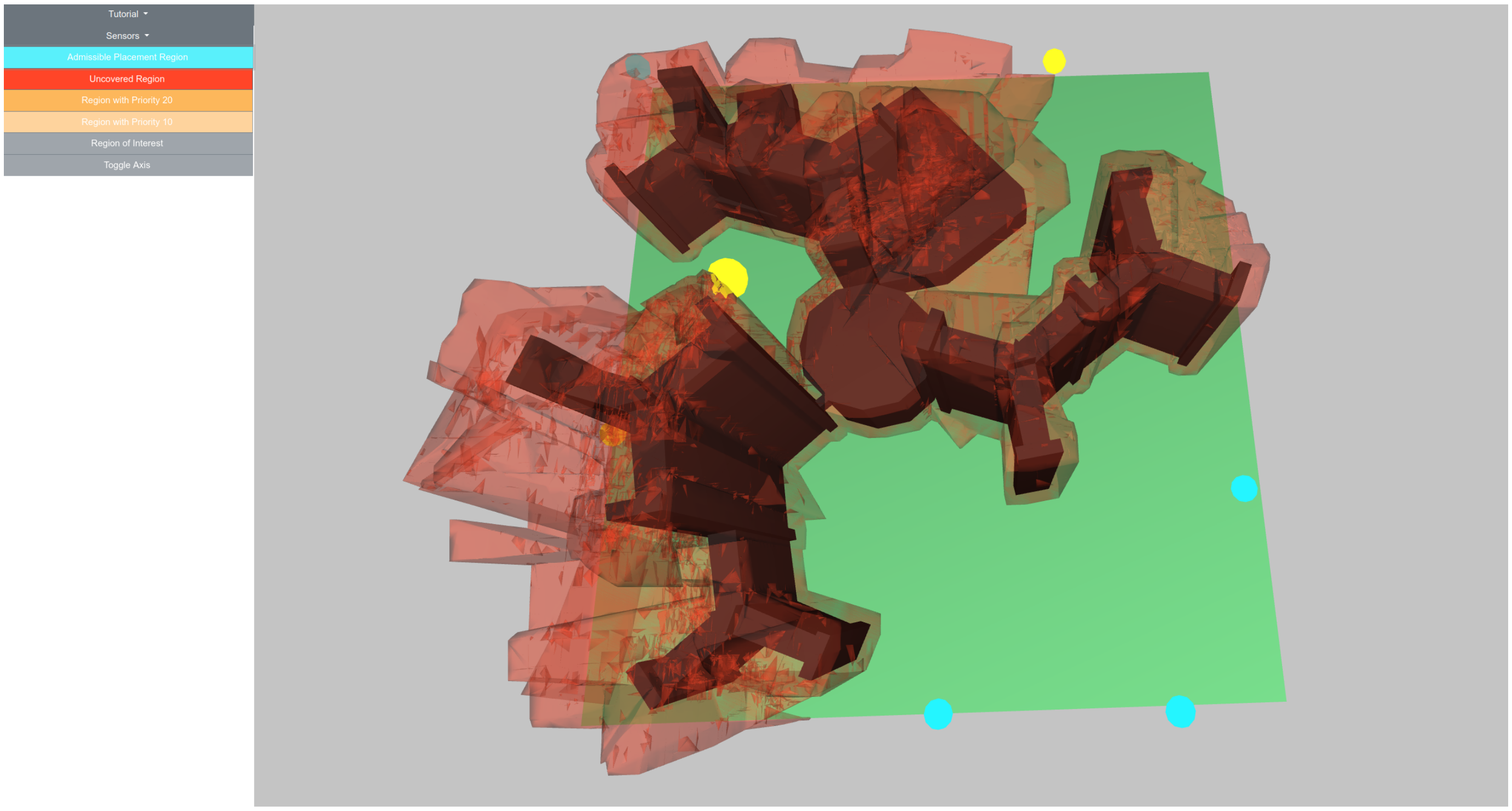

Figure 17: Uncovered region in VIC, seen from the above.

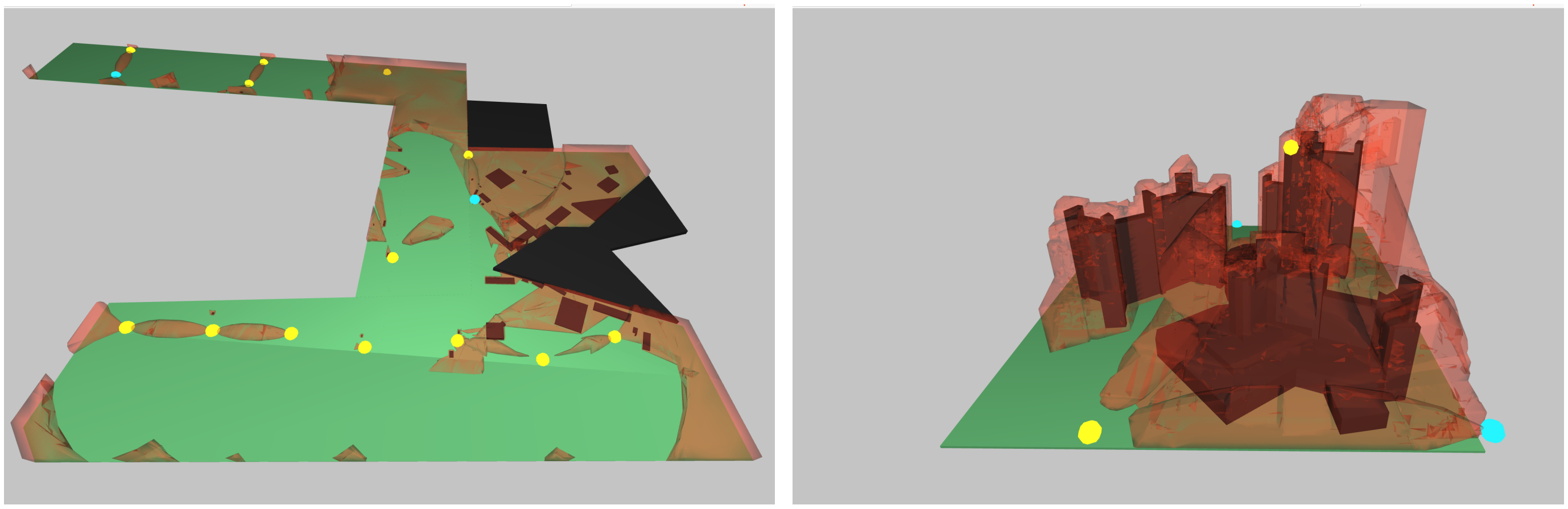

FCO VIC

Figure 18: Regions uncovered by the optimal deployment in the worst case.